\documentclass[10pt,journal,compsoc]{IEEEtran}
\usepackage{graphicx}
\usepackage[compress]{cite} 
\usepackage{subfig}
\usepackage{hyperref}
\hypersetup{
    bookmarks=true,         
    unicode=false,          
    pdftoolbar=true,        
    pdfmenubar=true,        
    pdffitwindow=false,     
    pdfstartview={FitH},    
    pdftitle={My title},    
    pdfauthor={Author},     
    pdfsubject={Subject},   
    pdfcreator={Creator},   
    pdfproducer={Producer}, 
    pdfkeywords={keyword1, key2, key3}, 
    pdfnewwindow=true,      
    colorlinks=true,       
    linkcolor=black,          
    citecolor=black,        
    filecolor=black,      
    urlcolor=black           
}
\ifCLASSOPTIONcompsoc
  \usepackage{cite}
\fi
\ifCLASSINFOpdf
\else
\fi
\begin{document}
%
\title{A Review on Generative Adversarial Networks: Algorithms, Theory, and Applications}
%
%
%
%

\author{Jie~Gui, Zhenan Sun, Yonggang Wen, Dacheng Tao, Jieping Ye

\thanks{J. Gui is with the Department of Computational Medicine and Bioinformatics,
University of Michigan, USA (e-mail: guijie@ustc.edu).}
\thanks{Z. Sun is with the Center for Research on Intelligent Perception and
Computing, Chinese Academy of Sciences, Beijing 100190, China (e-mail:
znsun@nlpr.ia.ac.cn).}
\thanks{Y. Wen is with the School of Computer Science and Engineering, Nanyang Technological University, Singapore 639798 (e-mail: ygwen@ntu.edu.sg).}
\thanks{D. Tao is with the UBTECH Sydney Artificial Intelligence Center and
the School of Information Technologies, the Faculty of Engineering and
Information Technologies, the University of Sydney, Australia. (e-mail:
Dacheng.Tao@Sydney.edu.au).}
\thanks{J. Ye is with DiDi AI Labs, P.R. China and University of Michigan, Ann Arbor.(e-mail:jpye@umich.edu).}
}

%
%

\markboth{Journal of \LaTeX\ Class Files,~Vol.~14, No.~8, August~2015}%
{Shell \MakeLowercase{\textit{et al.}}: Bare Demo of IEEEtran.cls for Computer Society Journals}
%



\IEEEtitleabstractindextext{%
\begin{abstract}
Generative adversarial networks (GANs) are a hot research topic recently. GANs have been widely studied since 2014, and a large number of algorithms have been proposed. However,
there is few comprehensive study explaining the connections among different GANs variants, and how they have evolved. In this paper, we attempt to provide a review on various GANs methods from the perspectives of algorithms, theory, and applications. Firstly, the motivations, mathematical representations, and structure of most GANs algorithms are introduced in details. Furthermore, GANs have been combined with other machine learning algorithms for specific applications, such as semi-supervised learning, transfer learning, and reinforcement learning. This paper compares the commonalities and differences of these GANs methods. Secondly, theoretical issues related to GANs are investigated. Thirdly, typical applications of GANs in image processing and computer vision, natural language processing, music, speech and audio, medical field, and data science are illustrated. Finally, the future open research problems for GANs are pointed out.
\end{abstract}

}

\maketitle

\IEEEdisplaynontitleabstractindextext

%
\IEEEpeerreviewmaketitle

\IEEEraisesectionheading{\section{Introduction}\label{sec:introduction}}

%
%
%
%
\IEEEPARstart{G}{enerative} adversarial networks (GANs) have become a hot research topic recently. Yann LeCun, a legend in deep learning, said in a Quora post ``GANs are the most interesting idea in the last 10 years in machine learning.'' There are a large number of papers related to GANs according to Google scholar. For example, there are about 11,800 papers related to GANs in 2018. That is to say, there are about 32 papers everyday and more than one paper every hour related to GANs in 2018.

GANs consist of two models: a generator and a discriminator. These two models are typically
implemented by neural networks, but they can be implemented with any form of differentiable system that maps data from one space to the other. The generator tries to capture the distribution of true examples for new data example generation. The discriminator is usually a binary
classifier, discriminating generated examples from the true examples as accurately as possible. The optimization of GANs is a minimax optimization problem. The optimization terminates at a saddle point that is a minimum with respect to the generator and a maximum with respect to the discriminator. That is, the optimization
goal is to reach Nash equilibrium \cite{ratliff2013characterization}. Then, the generator can be thought to have captured the real distribution of true examples.

Some previous work has adopted the concept of making two neural networks compete with each other. The most relevant work is predictability minimization \cite{schmidhuber1992learning}. The connections between predictability minimization and GANs can be found in \cite{goodfellow2014generative,schmidhuber2019unsupervised}.

The popularity and importance of GANs have led to several previous reviews. The difference from previous work is summarized in the following.
\begin{enumerate}
\item	\textit{GANs for specific applications}: There are surveys of using GANs for specific applications such as image synthesis and editing \cite{wu2017survey}, audio enhancement and synthesis \cite{Norberto2019Audio}. 

\item	\textit{General survey}: The earliest relevant review was probably the paper by Wang et al. \cite{wang2017generative} which mainly introduced the progress of GANs before 2017. References \cite{hong2019generative,creswell2018generative} mainly introduced the progress of GANs prior to 2018. The reference \cite{wang2019generative} introduced the architecture-variants and loss-variants of GANs only related to computer vision. Other related work can be found in \cite{hitawala2018comparative,zamorski2019generative,pan2019recent}. 
\end{enumerate}
As far as we know, this paper is the first to provide a comprehensive survey on GANs from the algorithm, theory, and application perspectives which introduces the latest progress. Furthermore, our paper focuses on applications not only to image processing and computer vision, but also sequential data such as natural language processing, and other related areas such as medical field.


The remainder of this paper is organized as follows: The related work is discussed in Section 2. Sections \ref{section:Algorithms}-\ref{section:Applications} introduce GANs from the algorithm, theory, and applications perspectives, respectively. Tables \ref{tab:1} and \ref{tab:2} show GANs' algorithms and applications which will be discussed in Sections \ref{section:Algorithms} and \ref{section:Applications}, respectively. The open research problems are discussed in Section 6 and Section 7 concludes the survey.
\begin{table*}
\caption{A overview of GANs' algorithms discussed in Section \ref{section:Algorithms}}\label{tab:1}
\begin{center}
\begin{tabular}{c c c}\hline
\multicolumn{2}{c}{GANs' Representative variants} & InfoGAN \cite{chen2016infogan}, cGANs \cite{mirza2014conditional}, CycleGAN \cite{lu2017conditional}, $f$-GAN \cite{nowozin2016f}, WGAN \cite{arjovsky2017wasserstein}, WGAN-GP \cite{gulrajani2017improved},\\
& & LS-GAN \cite{qi2017loss}\\\cline{1-3}
 & Objective function & LSGANs \cite{mao2017least,mao2019effectiveness}, hinge loss based GAN \cite{miyato2018spectral,lim2017geometric,tran2017deep}, MDGAN \cite{che2017mode}, unrolled GAN \cite{metz2016unrolled}, \\
 & & SN-GANs \cite{miyato2018spectral}, RGANs \cite{jolicoeur2018relativistic}  \\\cline{2-3}
& Skills & ImprovedGANs \cite{salimans2016improved}, AC-GAN \cite{odena2017conditional} \\\cline{2-3}
 &  & LAPGAN \cite{denton2015deep}, DCGANs \cite{radford2015unsupervised}, PGGAN \cite{karras2017progressive}, StackedGAN \cite{zhang2017stackgan}, SAGAN \cite{zhang2018self}, BigGANs \cite{brock2018large}, \\ 
 GANs training&  & StyleGAN \cite{karras2019style}, hybrids of autoencoders and GANs (EBGAN \cite{zhao2017energy}, \\
 &Structure & BEGAN \cite{berthelot2017began}, BiGAN \cite{donahue2016adversarial}/ALI \cite{dumoulin2016adversarially}, AGE \cite{ulyanov2018takes}),\\
 & & multi-discriminator learning (D2GAN \cite{nguyen2017dual}, GMAN \cite{durugkar2016generative}), \\
 & &  multi-generator learning (MGAN \cite{hoang2017multi}, MAD-GAN \cite{ghosh2018multi}), \\
 & & multi-GAN learning (CoGAN \cite{liu2016coupled}) \\\cline{1-3}
 & Semi-supervised learning & CatGANs \cite{springenberg2015unsupervised}, feature matching GANs \cite{salimans2016improved}, VAT \cite{miyato2018virtual}, $\Delta$-GAN \cite{gan2017triangle}, Triple-GAN \cite{chongxuan2017triple}\\\cline{2-3}
  &  & DANN \cite{ajakan2014domain}, CycleGAN \cite{zhu2017unpaired}, DiscoGAN \cite{kim2017learning}, DualGAN \cite{yi2017dualgan}, StarGAN \cite{choi2018stargan},  \\
Task driven GANs & Transfer learning & CyCADA \cite{hoffman2018cycada}, ADDA \cite{tzeng2017adversarial,wang2017catgan}, FCAN \cite{zhang2018fully}, \\
 & &  unsupervised pixel-level domain adaptation (PixelDA) \cite{bousmalis2017unsupervised} \\\cline{2-3}
 & Reinforcement learning & GAIL \cite{song2018multi}\\\hline
\end{tabular}
\end{center}
\end{table*}

\begin{table*}
\caption{Applications of GANs discussed in Section \ref{section:Applications}}\label{tab:2}
\begin{center}
\begin{tabular}{c c c}\hline
Field & Subfield & Method\\\cline{1-3}
 & Super-resolution & SRGAN \cite{ledig2017photo}, ESRGAN \cite{wang2018esrgan}, Cycle-in-Cycle GANs \cite{yuan2018unsupervised},  \\
& & SRDGAN \cite{guan2019srdgan}, TGAN \cite{ding2019tgan} \\\cline{2-3}
&  & DR-GAN \cite{tran2018representation}, TP-GAN \cite{huang2017beyond}, PG$^2$ \cite{ma2017pose}, PSGAN \cite{Jiang2019PSGAN}, \\
 &Image synthesis and manipulation &APDrawingGAN \cite{yi2019apdrawinggan}, IGAN \cite{zhu2016generative}, \\
Image processing and computer vision &  & introspective adversarial networks \cite{brock2016neural}, GauGAN \cite{park2019semantic} \\\cline{2-3}
& Texture synthesis & MGAN \cite{li2016precomputed}, SGAN \cite{jetchev2016texture}, PSGAN \cite{bergmann2017learning}\\\cline{2-3}
& Object detection & Segan \cite{ehsani2018segan}, perceptual GAN \cite{li2017perceptual}, MTGAN \cite{bai2018sod} \\\cline{2-3}
& Video & VGAN \cite{vondrick2016generating}, DRNET \cite{denton2017unsupervised}, Pose-GAN \cite{walker2017pose}, video2video \cite{wang2018video}, \\
& & MoCoGan \cite{tulyakov2018mocogan} \\\cline{1-3}
 & Natural language processing (NLP) & RankGAN \cite{lin2017adversarial}, IRGAN \cite{wang2017irgan,lu2019psgan}, TAC-GAN \cite{qiao2019mirrorgan}\\\cline{2-3}
Sequential data & Music & RNN-GAN (C-RNN-GAN) \cite{mogren2016c}, ORGAN \cite{guimaraes2017objective}, \\
& & SeqGAN \cite{lee2017seqgan,yu2017seqgan} \\\hline
\end{tabular}
\end{center}
\end{table*}

\section{Related work}
GANs belong to generative algorithms. Generative algorithms and discriminative algorithms are two categories of machine learning algorithms. If a machine learning algorithm is based on a fully probabilistic model of the observed data, this algorithm is generative. Generative algorithms have become more popular and important due to their wide practical applications. 

\subsection{Generative algorithms}
Generative algorithms can be classified into two classes: explicit density model and implicit density model.

\subsubsection{Explicit density model}
An explicit density model assumes the distribution and utilizes true data to train the model containing the distribution or fit the distribution parameters. When finished, new examples are produced utilizing the learned model or distribution. The explicit density models include maximum likelihood estimation (MLE), approximate
inference \cite{kingma2013auto,rezende2014stochastic}, and Markov chain method \cite{hinton1984boltzmann,ackley1985learning,hinton2006fast}. These explicit density models have an explicit distribution, but have limitations. For instance, MLE is conducted on true data and the parameters are updated directly based on the true data, which leads to an overly smooth generative model. The generative model learned by approximate inference can only approach the lower bound of the objective function rather than directly approach the objective function, because of the difficulty in solving the objective function. The Markov chain algorithm can be used to train generative models, but it is computationally expensive. Furthermore, the explicit density model has the problem of computational tractability. It may fail to represent the complexity of true data distribution and learn the high-dimensional data distributions \cite{nguyen2016synthesizing}.

\subsubsection{Implicit density model}
An implicit density model does not directly estimate or fit the data distribution. It produces data instances from the distribution without an explicit hypothesis \cite{bengio2014deep} and utilizes the produced examples to modify the model. Prior to GANs, the implicit density model generally needs to be trained utilizing either ancestral
sampling \cite{bengio2013generalized} or Markov chain-based sampling, which is inefficient and limits their practical applications. GANs belong to
the directed implicit density model category. A detailed summary and relevant papers can be found in \cite{goodfellow2016nips}.

\subsubsection{The comparison between GANs and other generative algorithms}
GANs were proposed to overcome the disadvantages of other generative algorithms. The basic idea behind adversarial learning is that the generator tries to create as realistic examples as possible to deceive the discriminator. The discriminator tries to distinguish fake examples from true examples. Both the generator and discriminator improve through adversarial learning. This adversarial process gives GANs notable advantages over other generative algorithms. More specifically, GANs have advantages over other generative algorithms as follows:
\begin{enumerate}
\item GANs can parallelize the generation, which is impossible for other generative algorithms such as PixelCNN \cite{salimans2017pixelcnn++} and fully visible belief networks (FVBNs) \cite{frey1996does,frey1998graphical}.
\item The generator design has few restrictions.
\item GANs are subjectively thought to produce better examples than other methods.
\end{enumerate}
Refer to \cite{goodfellow2016nips} for more detailed discussions about this.

\subsection{Adversarial idea}
The adversarial idea has been
successfully applied to many areas such as machine learning, artificial intelligence, computer vision and natural language processing.
The recent event that AlphaGo \cite{silver2016mastering} defeats world's top human player engages public interest
in artificial intelligence. The intermediate version of AlphaGo utilizes two networks
competing with each other.

Adversarial examples \cite{eykholt2018robust,kurakin2016adversarial,elsayed2018adversarial,jia2019comdefend,athalye2018obfuscated,zugner2018adversarial,dong2018boosting,szegedy2013intriguing,goodfellow2014explaining,kos2018adversarial} have the adversarial idea, too. Adversarial examples are those examples which are very different from the real examples, but are classified into a real category very
confidently, or those that are slightly different than the real examples, but are classified into a wrong category. This is a very hot research topic recently \cite{athalye2018obfuscated,zugner2018adversarial}. To be against adversarial attacks \cite{samangouei2018defense,akhtar2018threat}, references \cite{shen2017ape,lee2017generative} utilize GANs to conduct the right defense.

Adversarial machine learning \cite{huang2011adversarial} is a minimax problem. The defender, who builds the classifier that we want to work correctly, is searching over the parameter space to find the parameters that reduce the cost of the classifier as much as possible. Simultaneously, the attacker is searching over the inputs of the model to maximize the cost.  

The adversarial idea exists in adversarial networks, adversarial learning, and adversarial examples. However, they have different objectives. 

\section{Algorithms}\label{section:Algorithms}
In this section, we first introduce the original GANs. Then, the representative variants, training, evaluation of GANs, and task-driven GANs are introduced.

\subsection{Generative Adversarial Nets (GANs)}\label{subsection:GANs}
The GANs framework is straightforward to implement when the models are both neural networks. In order to learn the generator's distribution ${p_g}$ over data $x$, a prior on input noise variables is defined as ${p_z}\left( z \right)$ \cite{goodfellow2014generative} and $z$ is the noise variable. Then, GANs represent a mapping from noise space to data space as $G\left( {z,{\theta _g}} \right)$, where $G$ is a differentiable function represented by a neural network with parameters ${\theta _g}$. Other than $G$, the other neural network $D\left( {x,{\theta _d}} \right)$ is also defined with parameters ${\theta _d}$ and the output of $D\left( x \right)$ is a single scalar. $D\left( x \right)$ denotes the probability that $x$ was from the data rather than the generator $G$. The discriminator $D$ is trained to maximize the probability of giving the correct label to both training data and fake samples generated from the generator $G$. $G$ is trained to minimize $\log \left( {1 - D\left( {G\left( z \right)} \right)} \right)$ simultaneously  .

\subsubsection{Objective function}
Different objective functions can be used in GANs.

\paragraph{Original minimax game}~{}\newline
The objective function of GANs \cite{goodfellow2014generative} is
\begin{eqnarray}\label{equ:1}
\begin{array}{l}
 \mathop {\min }\limits_G {\rm{ }}\mathop {\max }\limits_D \;V\left( {D,G} \right) = {E_{x \sim {p_{data}}\left( x \right)}}\left[ {\log D\left( x \right)} \right] \\ 
 \quad  + {E_{z \sim {p_z}\left( z \right)}}\left[ {\log \left( {1 - D\left( {G\left( z \right)} \right)} \right)} \right]. \\ 
 \end{array}
\end{eqnarray}
$\log D\left( x \right)$ is the cross-entropy between ${\left[ {1\quad 0} \right]^T}$ 
and 
${\left[ {D\left( x \right)\quad 1 - D\left( x \right)} \right]^T}$. Similarly, $\log \left( {1 - D\left( {G\left( z \right)} \right)} \right)$ is the cross-entropy between ${\left[ {0\quad 1} \right]^T}$ 
and 
${\left[ {D\left( {G\left( z \right)} \right)\quad 1 - D\left( {G\left( z \right)} \right)} \right]^T}$. 
For fixed $G$, the optimal discriminator $D$ is given by \cite{goodfellow2014generative}:
\begin{eqnarray}\label{equ:2}
D_G^*\left( x \right) = \frac{{{p_{data}}\left( x \right)}}{{{p_{data}}\left( x \right) + {p_g}\left( x \right)}}.
\end{eqnarray}
The minmax game in (\ref{equ:1}) can be reformulated as:
\begin{eqnarray}\label{equ:3}
\begin{array}{l}
 C(G) = \mathop {\max }\limits_D \;V\left( {D,G} \right) \\ 
  = {E_{x \sim {p_{data}}}}\left[ {\log D_G^*\left( x \right)} \right] \\ 
 \quad \quad  + {E_{z \sim {p_z}}}\left[ {\log \left( {1 - D_G^*\left( {G\left( z \right)} \right)} \right)} \right] \\ 
  = {E_{x \sim {p_{data}}}}\left[ {\log D_G^*\left( x \right)} \right] + {E_{x \sim {p_g}}}\left[ {\log \left( {1 - D_G^*\left( x \right)} \right)} \right] \\ 
  = {E_{x \sim {p_{data}}}}\left[ {\log \frac{{{p_{data}}\left( x \right)}}{{\frac{1}{2}\left( {{p_{data}}\left( x \right) + {p_g}\left( x \right)} \right)}}} \right] \\ 
 \quad \quad  + {E_{x \sim {p_g}}}\left[ {\frac{{{p_g}\left( x \right)}}{{\frac{1}{2}\left( {{p_{data}}\left( x \right) + {p_g}\left( x \right)} \right)}}} \right] - 2\log 2. \\ 
 \end{array}
\end{eqnarray}
The definition of Kullback–Leibler (KL) divergence and Jensen-Shannon (JS) divergence between two probabilistic distributions $p\left( x \right)$ and $q\left( x \right)$ are defined as \begin{eqnarray}\label{equ:4}
KL(\left. p \right\|q) = \int {p\left( x \right)\log \frac{{p\left( x \right)}}{{q\left( x \right)}}} dx,
\end{eqnarray}

\begin{eqnarray}\label{equ:5}
JS(\left. p \right\|q) = \frac{1}{2}KL(\left. p \right\|\frac{{p + q}}{2}) + \frac{1}{2}KL(\left. q \right\|\frac{{p + q}}{2}).
\end{eqnarray}
Therefore, (\ref{equ:3}) is equal to
\begin{eqnarray}\label{equ:6}
\begin{array}{l}
 C(G) = KL(\left. {{p_{data}}} \right\|\frac{{{p_{data}} + {p_g}}}{2}) + KL(\left. {{p_g}} \right\|\frac{{{p_{data}} + {p_g}}}{2}) \\ 
 \quad \quad \quad  - 2\log 2 \\ 
  = 2JS(\left. {{p_{data}}} \right\|{p_g}) - 2\log 2. \\ 
 \end{array}
 \end{eqnarray}
 Thus, the objective function of GANs is related to both KL divergence and JS divergence.

\paragraph{Non-saturating game}~{}\newline
It is possible that the Equation (\ref{equ:1}) cannot provide sufficient gradient for $G$  to learn well in practice. Generally speaking, $G$ is poor in early learning and samples are clearly different from the training data. Therefore, $D$ can reject the generated samples with high confidence. In this situation, $\log \left( {1 - D\left( {G\left( z \right)} \right)} \right)$ saturates. We can train $G$ to maximize $\log \left( {D\left( {G\left( z \right)} \right)} \right)$ rather than minimize $\log \left( {1 - D\left( {G\left( z \right)} \right)} \right)$. The cost for the generator then becomes
\begin{eqnarray}\label{equ:7}
\begin{array}{l}
 {J^{(G)}} = {E_{z \sim {p_z}\left( z \right)}}\left[ { - \log \left( {D\left( {G\left( z \right)} \right)} \right)} \right] \\ 
  = {E_{x \sim {p_g}}}\left[ { - \log \left( {D\left( x \right)} \right)} \right]. \\ 
 \end{array}
 \end{eqnarray}
This new objective function results in the same fixed point of the dynamics of $D$ and $G$ but provides much larger gradients early in learning. The non-saturating game is heuristic, not being motivated by theory. However, the non-saturating game has other problems such as unstable numerical gradient for training $G$. With optimal $D_G^*$, we have
\begin{eqnarray}\label{equ:8}
\begin{array}{l}
 {E_{x \sim {p_g}}}\left[ { - \log \left( {D_G^*\left( x \right)} \right)} \right] + {E_{x \sim {p_g}}}\left[ {\log \left( {1 - D_G^*\left( x \right)} \right)} \right] \\ 
  = {E_{x \sim {p_g}}}\left[ {\log \frac{{\left( {1 - D_G^*\left( x \right)} \right)}}{{D_G^*\left( x \right)}}} \right] = {E_{x \sim {p_g}}}\left[ {\log \frac{{{p_g}\left( x \right)}}{{{p_{data}}\left( x \right)}}} \right] \\ 
  = KL(\left. {{p_g}} \right\|{p_{data}}). \\ 
 \end{array}
 \end{eqnarray}
 Therefore, ${E_{x \sim {p_g}}}\left[ { - \log \left( {D_G^*\left( x \right)} \right)} \right]$ is equal to
\begin{eqnarray}\label{equ:9}
\begin{array}{l}
 {E_{x \sim {p_g}}}\left[ { - \log \left( {D_G^*\left( x \right)} \right)} \right] \\ 
  = KL(\left. {{p_g}} \right\|{p_{data}}) - {E_{x \sim {p_g}}}\left[ {\log \left( {1 - D_G^*\left( x \right)} \right)} \right]. \\ 
 \end{array}
 \end{eqnarray}
 From (\ref{equ:3}) and (\ref{equ:6}), we have
\begin{eqnarray}\label{equ:10}
\begin{array}{l}
 {E_{x \sim {p_{data}}}}\left[ {\log D_G^*\left( x \right)} \right] + {E_{x \sim {p_g}}}\left[ {\log \left( {1 - D_G^*\left( x \right)} \right)} \right] \\ 
  = 2JS(\left. {{p_{data}}} \right\|{p_g}) - 2\log 2. \\ 
 \end{array}
\end{eqnarray}
 Therefore, ${E_{x \sim {p_g}}}\left[ {\log \left( {1 - D_G^*\left( x \right)} \right)} \right]$ equals
 \begin{eqnarray}\label{equ:11}
 \begin{array}{l}
 {E_{x \sim {p_g}}}\left[ {\log \left( {1 - D_G^*\left( x \right)} \right)} \right] \\ 
  = 2JS(\left. {{p_{data}}} \right\|{p_g}) - 2\log 2 - {E_{x \sim {p_{data}}}}\left[ {\log D_G^*\left( x \right)} \right]. \\ 
 \end{array}
 \end{eqnarray}
 By substituting (\ref{equ:11}) into (\ref{equ:9}), (\ref{equ:9}) reduces to
 \begin{eqnarray}\label{equ:12}
 \begin{array}{l}
 {E_{x \sim {p_g}}}\left[ { - \log \left( {D_G^*\left( x \right)} \right)} \right] \\ 
  = KL(\left. {{p_g}} \right\|{p_{data}}) - 2JS(\left. {{p_{data}}} \right\|{p_g}) +  \\ 
 {E_{x \sim {p_{data}}}}\left[ {\log D_G^*\left( x \right)} \right] + 2\log 2. \\ 
 \end{array}
 \end{eqnarray}
 From (\ref{equ:12}), we can see that the optimization of the alternative $G$ loss in the non-saturating game is contradictory because the first term aims to make the divergence between the generated distribution and the real distribution as small as possible while the second term aims to make the divergence between these two distributions as large as possible due to the negative sign. This will bring unstable numerical gradient for training $G$. Furthermore, KL divergence is not a symmetrical quantity, which is reflected from the following two examples
\begin{itemize}
\item If ${p_{data}}\left( x \right) \to 0$ and ${p_g}\left( x \right) \to 1$, we have $KL(\left. {{p_g}} \right\|{p_{data}}) \to  + \infty $.
\item If ${p_{data}}\left( x \right) \to 1$ and ${p_g}\left( x \right) \to 0$, we have $KL(\left. {{p_g}} \right\|{p_{data}}) \to 0$.
\end{itemize}
The penalizations for two errors made by $G$ are completely different. The first error is that $G$ produces implausible samples and the penalization is rather large. The second error is that $G$ does not produce real samples and the penalization is quite small. The first error is that the generated samples are inaccurate while the second error is that generated samples are not diverse enough. Based on this, $G$ prefers producing repeated but safe samples rather than taking risk to produce different but unsafe samples, which has the mode collapse problem.

\begin{figure}
\begin{center}
\scalebox{0.48}{\includegraphics{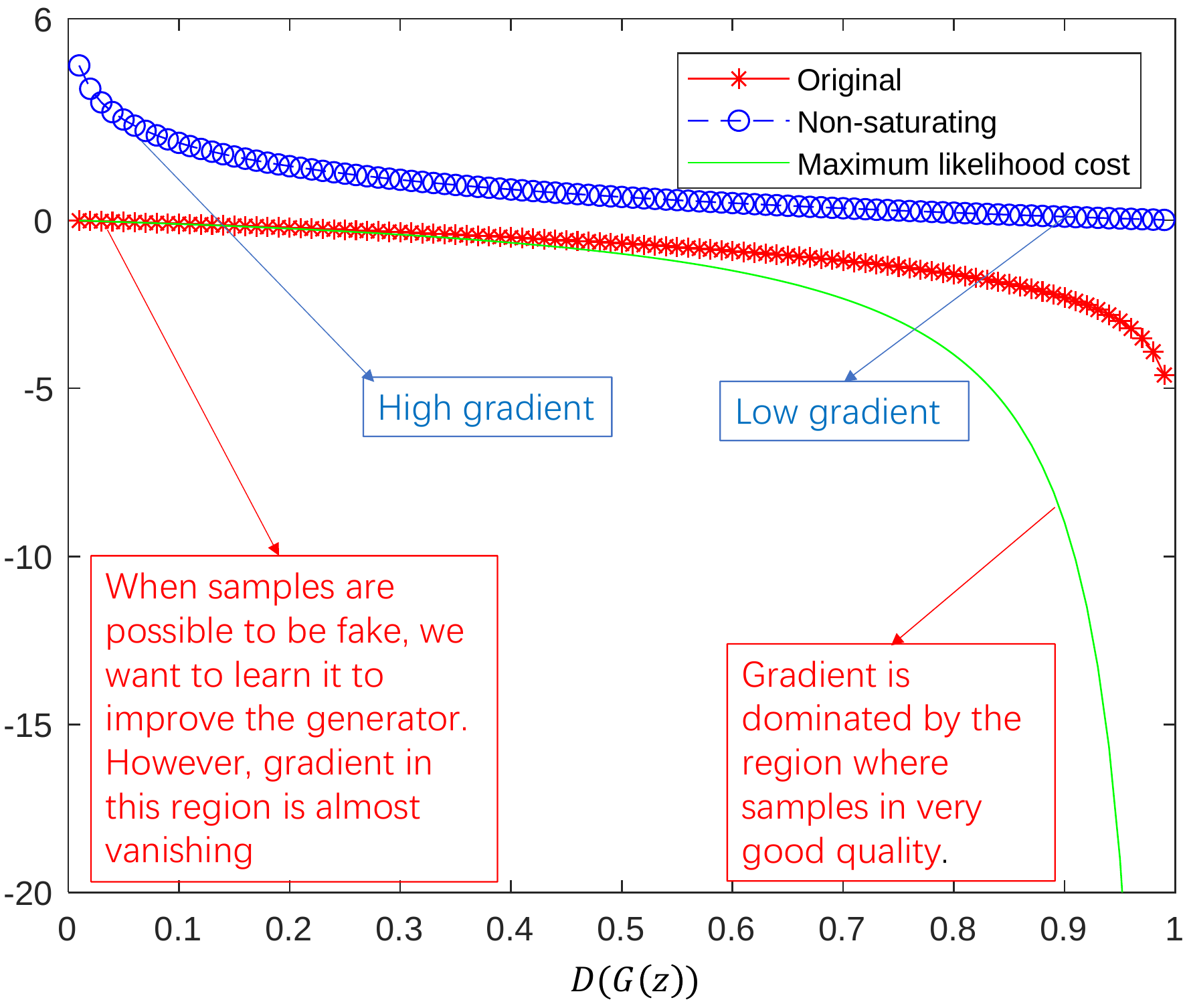}}
\end{center}
   \caption{The three curves of ``Original'', ``Non-saturating'', and ``Maximum likelihood cost'' denotes $\log \left( {1 - D\left( {G\left( z \right)} \right)} \right)$, $ - \log \left( {D\left( {G\left( z \right)} \right)} \right)$, and $ -D(G(z))/(1-D(G(z)))$ in (\ref{equ:1}), (\ref{equ:7}), and (\ref{equ:13}), respectively. The cost that the generator has for generating a sample $G(z)$ is only decided by the discriminator's response to that generated sample. The larger probability the discriminator gives the real label to the generated sample, the less cost the generator gets. This figure is reproduced from \cite{goodfellow2016nips,goodfellow2014distinguishability}.}
\label{fig:2}
\end{figure}

\paragraph{Maximum likelihood game}~{}\newline
There are many methods to approximate (\ref{equ:1}) in GANs. Under the assumption that the discriminator is optimal, minimizing
 \begin{eqnarray}\label{equ:13}
\begin{array}{l}
 {J^{(G)}} = {E_{z \sim {p_z}\left( z \right)}}\left[ { - \exp \left( {{\sigma ^{ - 1}}\left( {D\left( {G\left( z \right)} \right)} \right)} \right)} \right] \\ 
  = {E_{z \sim {p_z}\left( z \right)}}\left[ { - {{D\left( {G\left( z \right)} \right)} \mathord{\left/
 {\vphantom {{D\left( {G\left( z \right)} \right)} {\left( {1 - D\left( {G\left( z \right)} \right)} \right)}}} \right.
 \kern-\nulldelimiterspace} {\left( {1 - D\left( {G\left( z \right)} \right)} \right)}}} \right], \\
 \end{array}
\end{eqnarray}
where $\sigma$ is the logistic sigmoid function, equals minimizing (\ref{equ:1}) \cite{goodfellow2014distinguishability}. The demonstration of this equivalence can be found in Section 8.3 of \cite{goodfellow2016nips}. Furthermore, there are other possible ways of approximating maximum likelihood within the GANs framework \cite{nowozin2016f}. A comparison of original zero-sum game, non-saturating game, and maximum likelihood game is shown in Fig. \ref{fig:2}.

Three observations can be obtained from Fig. \ref{fig:2}.
\begin{itemize}
\item First, when the sample is possible to be fake, that is on the left end of the figure, both the maximum likelihood game and the original minimax game suffer from gradient vanishing. The heuristically motivated non-saturating game does not have this problem. 
\item Second, maximum likelihood game also has the problem that almost all of the gradient is from the right end of the curve, which means that a rather small number of samples in each minibatch dominate the gradient computation. This demonstrates that variance reduction methods could be an important research direction for improving the performance of GANs based on maximum likelihood game.
\item Third, the heuristically motivated non-saturating game has lower sample variance, which is the possible reason that it is more successful in real applications.
\end{itemize}
GAN Lab \cite{kahng2018gan} is proposed as the interactive visualization tool designed for non-experts to learn and experiment with GANs. Bau et al. \cite{bau2018gan} present an analytic framework to visualize and understand GANs. 

\subsection{GANs' representative variants}
There are many papers related to GANs \cite{kocaoglu2017causalgan,feizi2017understanding,farnia2018convex,zhao2018understanding,jahanian2019steerability,zhu2019deconstructing} such as CSGAN \cite{kancharagunta2019csgan} and LOGAN \cite{Wu2019LOGAN}. In this subsection, we will introduce GANs' representative variants.

\subsubsection{InfoGAN}
Rather than utilizing a single unstructured noise vector $z$, InfoGAN \cite{chen2016infogan} proposes to decompose the input noise vector into two parts: $z$, which is seen as incompressible noise; $c$, which is called the latent code and will target the significant structured semantic features of the real data distribution. InfoGAN \cite{chen2016infogan} aims to solve 
\begin{eqnarray}\label{equ:14}
\mathop {\min }\limits_G {\rm{ }}\mathop {\max }\limits_D \;{V_I}\left( {D,G} \right) = V\left( {D,G} \right) - \lambda I(c;G(z,c)),
\end{eqnarray}
where $V\left( {D,G} \right)$ is the objective function of original GAN, $G(z,c)$ is the generated sample, $I$ is the mutual information, and $\lambda$ is the tunable regularization parameter. Maximizing $I(c;G(z,c))$ means maximizing the mutual information between $c$ and $G(z,c)$ to make $c$ contain as much important and meaningful features of the real samples as possible. However, $I(c;G(z,c))$ is difficult to optimize directly in practice since it requires access to the posterior $P(c|x)$. Fortunately, we can have a lower bound of $I(c;G(z,c))$ by defining an auxiliary distribution $Q(c|x)$ to approximate $P(c|x)$. The final objective function of InfoGAN \cite{chen2016infogan} is
\begin{eqnarray}\label{equ:15}
\mathop {\min }\limits_G {\rm{ }}\mathop {\max }\limits_D \;{V_I}\left( {D,G} \right) = V\left( {D,G} \right) - \lambda {L_I}(c;Q),
\end{eqnarray}
where ${L_I}(c;Q)$ is the lower bound of $I(c;G(z,c))$. InfoGAN has several variants such as causal InfoGAN \cite{kurutach2018learning} and semi-supervised InfoGAN (ss-InfoGAN) \cite{spurr2017guiding}. 

\subsubsection{Conditional GANs (cGANs)}
GANs can be extended to a conditional model if both the discriminator and generator are conditioned on some extra information $y$. The objective function of conditional GANs \cite{mirza2014conditional} is:
\begin{eqnarray}\label{equ:16}
\begin{array}{l}
 \mathop {\min }\limits_G {\rm{ }}\mathop {\max }\limits_D \;V\left( {D,G} \right) = {E_{x \sim {p_{data}}\left( x \right)}}\left[ {\log D\left( {\left. x \right|y} \right)} \right] \\ 
 \quad \quad \quad  + {E_{z \sim {p_z}\left( z \right)}}\left[ {\log \left( {1 - D\left( {G\left( {\left. z \right|y} \right)} \right)} \right)} \right]. \\ 
 \end{array}
 \end{eqnarray}
By comparing (\ref{equ:15}) and (\ref{equ:16}), we can see that the generator of InfoGAN is similar to that of cGANs. However, the latent code $c$ of InfoGAN is not known, and it is discovered by training. Furthermore, InfoGAN has an additional network $Q$ to output the conditional variables $Q(c|x)$.
 
Based on cGANs, we can generate samples conditioning on class labels \cite{odena2017conditional,nguyen2017plug}, text \cite{reed2016generative,hong2018inferring,zhang2017stackgan}, bounding box and keypoints \cite{reed2016learning}. In \cite{zhang2017stackgan,zhang2019stackgan++}, text to photo-realistic image synthesis is conducted with stacked generative adversarial networks (SGAN) \cite{huang2017stacked}. cGANs have been used for convolutional face generation \cite{gauthier2014conditional}, face aging \cite{antipov2017face}, image translation \cite{tang2019multi}, synthesizing outdoor
images having specific scenery attributes \cite{karacan2016learning}, natural image description \cite{dai2017towards}, and 3D-aware scene manipulation \cite{yao20183d}. Chrysos et al. \cite{chrysos2018robust} proposed robust cGANs. Thekumparampil et al. \cite{thekumparampil2018robustness} discussed the robustness of conditional GANs to noisy labels. Conditional CycleGAN \cite{lu2017conditional} uses cGANs with cyclic consistency. Mode seeking GANs (MSGANs) \cite{mao2019mode} proposes a simple yet effective regularization term to address the mode collapse issue for cGANs.

The discriminator of original GANs \cite{goodfellow2014generative} is trained to maximize the log-likelihood that it assigns to the correct source \cite{odena2017conditional}:
\begin{eqnarray}\label{equ:17}
\begin{array}{l}
 L = E\left[ {\log P\left( {\left. {S = real} \right|{X_{real}}} \right)} \right] \\ 
 \quad  + E\left[ {\log \left( {P\left( {\left. {S = fake} \right|{X_{fake}}} \right)} \right)} \right], \\ 
 \end{array}
 \end{eqnarray}
 which is equal to (\ref{equ:1}). The objective function of the auxiliary classifier GAN (AC-GAN) \cite{odena2017conditional} has two parts: the loglikelihood of the correct source, ${L_S}$, and the loglikelihood of the correct class label, ${L_C}$. ${L_S}$ is equivalent to $L$ in (\ref{equ:17}). ${L_C}$ is defined as
 \begin{eqnarray}\label{equ:18}
\begin{array}{l}
 {L_C} = E\left[ {\log P\left( {\left. {C = c} \right|{X_{real}}} \right)} \right] \\ 
 \quad  + E\left[ {\log \left( {P\left( {\left. {C = c} \right|{X_{fake}}} \right)} \right)} \right]. \\
 \end{array}
 \end{eqnarray}
The discriminator and generator of AC-GAN is to maximize ${L_C} + {L_S}$ and ${L_C} - {L_S}$, respectively. AC-GAN was the first variant of GANs that was able to produce recognizable examples of all the ImageNet \cite{deng2009imagenet} classes. 

Discriminators of most cGANs based methods \cite{denton2015deep,perarnau2016invertible,saito2017temporal,dumoulin2016adversarially,sricharan2017semi} feed conditional information $y$ into the discriminator by simply concatenating (embedded) $y$ to the input or to the feature vector at some middle layer. cGANs with projection discriminator \cite{miyato2018cgans} adopts an inner product between the condition vector $y$ and the feature vector.

Isola et al. \cite{isola2017image} used cGANs and sparse regularization for image-to-image translation. The corresponding software is called pix2pix. In GANs, the generator learns a mapping from random noise $z$ to $G\left( z \right)$. In contrast, there is no noise input in the generator of pix2pix. A novelty of pix2pix is that the generator of pix2pix learns a mapping from an observed image $y$  to output image $G\left( y \right)$, for example, from a grayscale image to a color image. The objective of cGANs in \cite{isola2017image} can be expressed as 
\begin{eqnarray}\label{equ:19}
\begin{array}{l}
 {L_{cGANs}}\left( {D,G} \right) = {E_{x,y}}\left[ {\log D\left( {x,y} \right)} \right] \\ 
 \quad  + {E_y}\left[ {\log \left( {1 - D\left( {y,G\left( y \right)} \right)} \right)} \right]. \\ 
 \end{array}
 \end{eqnarray}
 
 Furthermore, ${l_1}$ distance is used:
 \begin{eqnarray}\label{equ:20}
 {L_{{l_1}}}\left( G \right) = {E_{x,y}}\left[ {{{\left\| {x - G(y)} \right\|}_1}} \right].
 \end{eqnarray}
The final objective of \cite{isola2017image} is
\begin{eqnarray}\label{equ:21}
{L_{cGANs}}\left( {D,G} \right) + \lambda {L_{{l_1}}}\left( G \right),
 \end{eqnarray}
where $\lambda$ is the free parameter. 
\begin{figure}
\begin{center}
\scalebox{0.4}{\includegraphics{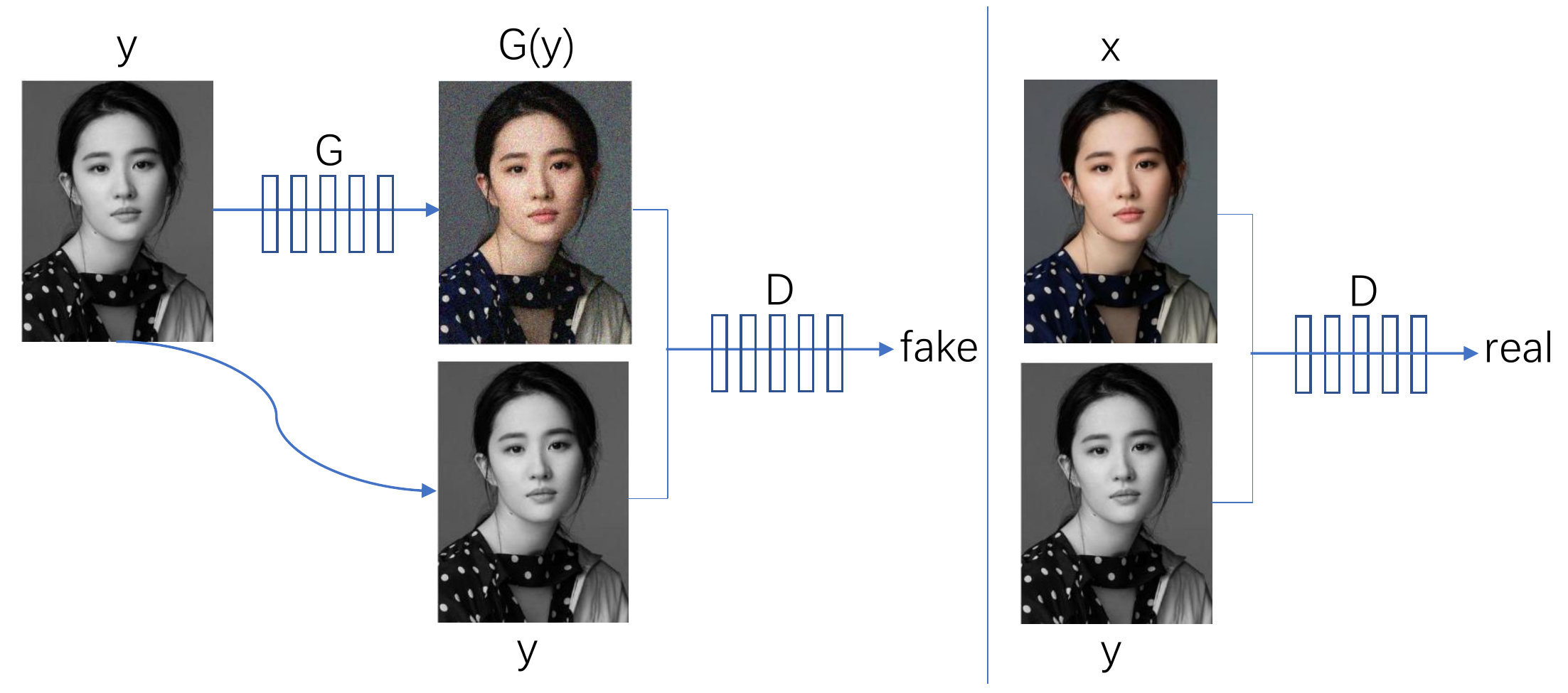}}
\end{center}
   \caption{Illustration of pix2pix: Training a conditional GANs to map grayscale$ \to $color. The discriminator $D$ learns to classify between real {grayscale, color} tuples and fake (synthesized by the generator). The generator $G$ learns to fool the discriminator. Different from the original GANs, both the generator and discriminator observe the input grayscale image and there is no noise input for the generator of pix2pix.}
\label{fig:3}
\end{figure}
As a follow-up to pix2pix, pix2pixHD \cite{wang2018high} used cGANs and feature matching loss for high-resolution image synthesis and semantic manipulation. With the discriminators, the learning problem is a multi-task learning problem:
\begin{eqnarray}\label{equ:22}
\mathop {\min }\limits_G {\rm{ }}\mathop {\max }\limits_{{D_1},{D_2},{D_3}} \;\sum\limits_{k = 1,2,3} {{L_{GAN}}\left( {G,{D_k}} \right)} .
 \end{eqnarray}
The training set is given as a set of pairs of corresponding images $\left\{ {\left( {{s_i},{x_i}} \right)} \right\}$, where ${x_i}$ is a natural photo and ${s_i}$ is a corresponding semantic label map. The $i$th-layer feature extractor of discriminator ${D_k}$ is denoted as $D_k^{(i)}$ (from input to the $i$th layer of ${D_k}$). The feature matching loss ${L_{FM}}\left( {G,{D_k}} \right)$ is:
\begin{eqnarray}\label{equ:23}
\begin{array}{l}
 {L_{FM}}\left( {G,{D_k}} \right) =  \\ 
 {E_{\left( {s,x} \right)}}\sum\limits_{i = 1}^T {\frac{1}{{{N_i}}}} \left[ {{{\left\| {D_k^{(i)}\left( {s,x} \right) - D_k^{(i)}\left( {s,G\left( s \right)} \right)} \right\|}_1}} \right], \\ 
 \end{array}
 \end{eqnarray}
where ${N_i}$ is the number of elements in each layer and $T$ denotes the total number of layers. The final objective function of \cite{wang2018high} is
\begin{eqnarray}\label{equ:24}
\mathop {\min }\limits_G {\rm{ }}\mathop {\max }\limits_{{D_1},{D_2},{D_3}} \;\sum\limits_{k = 1,2,3} {\left( {{L_{GAN}}\left( {G,{D_k}} \right) + \lambda {L_{FM}}\left( {G,{D_k}} \right)} \right)} .
 \end{eqnarray}

\subsubsection{CycleGAN}
Image-to-image translation is a class of graphics and vision problems where the goal is to learn the mapping between an output image and an input image using a training set of aligned image pairs. When paired training data is available, reference \cite{isola2017image} can be used for these image-to-image translation tasks. However, reference \cite{isola2017image} can not be used for unpaired data (no input/output pairs), which was well solved by Cycle-consistent GANs (CycleGAN) \cite{zhu2017unpaired}. CycleGAN is an important progress for unpaired data. It is proved that cycle-consistency is an upper bound of the conditional entropy \cite{li2017alice}. CycleGAN can be derived as a special case within the proposed variational inference (VI) framework \cite{tiao2018cycle}, naturally establishing its relationship with approximate Bayesian inference methods. 

The basic idea of DiscoGAN \cite{kim2017learning} and CycleGAN \cite{zhu2017unpaired} is nearly the same. Both of them were proposed separately nearly at the same time. The only difference between CycleGAN \cite{zhu2017unpaired} and DualGAN \cite{yi2017dualgan} is that DualGAN uses the loss format advocated by Wasserstein GAN (WGAN) rather than the sigmoid cross-entropy loss used in CycleGAN.

\subsubsection{$f$-GAN}
As we know, Kullback-Leibler (KL) divergence measures the difference between two given probability distributions. A large class of assorted divergences are the so called Ali-Silvey distances, also known as the $f$-divergences \cite{csiszar2004information}. Given two probability distributions $P$ and $Q$ which have, respectively, an absolutely continuous density function $p$ and $q$ with regard to a base measure $dx$  defined on the domain $X$, the $f$-divergence is defined,
\begin{eqnarray}\label{equ:25}
{D_f}\left( {P\left\| Q \right.} \right) = \int\limits_X {q\left( x \right)} f\left( {\frac{{p\left( x \right)}}{{q\left( x \right)}}} \right)dx.
\end{eqnarray}
Different choices of $f$ recover popular divergences as special cases of $f$-divergence. For example, if $f\left( a \right) = a\log a$, $f$-divergence becomes KL divergence. The original GANs \cite{goodfellow2014generative} is a special case of $f$-GAN \cite{nowozin2016f} which is based on $f$-divergence. The reference \cite{nowozin2016f} shows that any $f$-divergence can be used for training GAN. Furthermore, the reference \cite{nowozin2016f} discusses the advantages of different choices of divergence functions on both the quality of the produced generative models and training complexity. Im et al. \cite{im2018quantitatively} quantitatively evaluated GANs with divergences proposed for training. Uehara et al. \cite{uehara2016generative} extend the $f$-GAN further, where the $f$-divergence is directly minimized in the generator step and the ratio of the distributions of real and generated data are predicted in the discriminator step.

\subsubsection{Integral Probability Metrics (IPMs)}
Denoting ${\cal P}$ the set of all Borel probability measures on a topological space $(M,{\cal A})$. The integral probability metric (IPM) \cite{sriperumbudur2010hilbert} between two probability distributions $P \in {\cal P}$ and $Q \in {\cal P}$ is defined as 
\begin{eqnarray}\label{equ:26}
{\gamma _{\cal F}}(P,Q) = \mathop {\sup }\limits_{f \in {\cal F}} \left| {\int_M {fdP - \int_M {fdQ} } } \right|,
\end{eqnarray}
where ${\cal F}$ is a class of real-valued bounded measurable functions on $M$. Nonparametric density estimation and convergence rates for GANs under Besov IPM Losses is discussed in \cite{NIPS2019_9109}. IPMs include such as RKHS-induced maximum mean discrepancy (MMD) as well as the Wasserstein distance used in Wasserstein GANs (WGAN).

\paragraph{Maximum Mean Discrepancy (MMD)}~{}\newline
The maximum mean discrepancy (MMD) \cite{gretton2012kernel} is a measure of the difference between two distributions $P$ and $Q$ given by the supremum over a function space ${\cal F}$ of differences between the expectations with regard to two distributions. The MMD is defined by:
\begin{eqnarray}\label{equ:27}
\begin{array}{l}
 MMD({\cal F},P,Q) =  \\ 
 \quad \mathop {\sup }\limits_{f \in {\cal F}} \left( {{E_{X \sim P}}\left[ {f\left( X \right)} \right] - {E_{Y \sim Q}}\left[ {f\left( Y \right)} \right]} \right). \\ 
 \end{array}
\end{eqnarray}
MMD has been used for deep generative models \cite{dziugaite2015training,li2017mmd,li2015generative} and model criticism \cite{sutherland2016generative}.

\paragraph{Wasserstein GAN (WGAN)}~{}\newline
WGAN \cite{arjovsky2017wasserstein} conducted a comprehensive theoretical analysis of how the Earth Mover (EM) distance behaves in comparison with popular probability distances and divergences such as the total variation (TV) distance, the Kullback-Leibler (KL) divergence, and the Jensen-Shannon (JS) divergence utilized in the context of learning distributions. The definition of the EM distance is 
\begin{eqnarray}\label{equ:28}
W({p_{data}},{p_g}) = \mathop {\inf }\limits_{\gamma  \in \Pi \left( {{p_{data}},{p_g}} \right)} {E_{(x,y) \in \gamma }}\left[ {\left\| {x - y} \right\|} \right],
\end{eqnarray}
where $\Pi \left( {{p_{data}},{p_g}} \right)$ denotes the set of all joint distributions $\gamma \left( {x,y} \right)$ whose marginals are ${p_{data}}$ and ${p_g}$, respectively. However, the infimum in (\ref{equ:28}) is highly intractable. The reference \cite{arjovsky2017wasserstein} uses the following equation to approximate the EM distance
\begin{eqnarray}\label{equ:29}
\mathop {\max }\limits_{w \in {\cal W}} {E_{x \sim {p_{data}}\left( x \right)}}\left[ {{f_w}\left( x \right)} \right] - {E_{z \sim {p_z}\left( z \right)}}\left[ {{f_w}\left( {G\left( z \right)} \right)} \right],
\end{eqnarray}
where there is a parameterized family of functions ${\{ {f_w}\} _{w \in {\cal W}}}$ that are all $K$-Lipschitz for some $K$ and ${{f_w}}$ can be realized by the discriminator $D$. When $D$ is optimized, (\ref{equ:29}) denotes the approximated EM distance. Then the aim of $G$ is to minimize (\ref{equ:29}) to make the generated distribution as close to the real distribution as possible. Therefore, the overall objective function of WGAN is
\begin{eqnarray}\label{equ:30}
\begin{array}{l}
 \mathop {\min }\limits_G \mathop {\max }\limits_{w \in {\cal W}} {E_{x \sim {p_{data}}\left( x \right)}}\left[ {{f_w}\left( x \right)} \right] - {E_{z \sim {p_z}\left( z \right)}}\left[ {{f_w}\left( {G\left( z \right)} \right)} \right] \\ 
  = \mathop {\min }\limits_G \mathop {\max }\limits_D {E_{x \sim {p_{data}}\left( x \right)}}\left[ {D\left( x \right)} \right] - {E_{z \sim {p_z}\left( z \right)}}\left[ {D\left( {G\left( z \right)} \right)} \right]. \\ 
 \end{array}
\end{eqnarray}
By comparing (\ref{equ:1}) and (\ref{equ:30}), we can see three differences between the objective function of original GANs and that of WGAN: 
\begin{itemize}
    \item First, there is no $log$ in the objective function of WGAN. 
    \item Second, the $D$ in original GANs is utilized as a binary classifier while $D$ utilized in WGAN is to approximate the Wasserstein distance, which is a regression task. Therefore, the sigmoid in the last layer of $D$ is not used in the WGAN. The output of the discriminator of the original GANs is between zero and one while there is no constraint for that of WGAN.
    \item Third, the $D$ in WGAN is required to be $K$-Lipschitz for some $K$ and therefore WGAN uses weight clipping.
\end{itemize}
Compared with traditional GANs training, WGAN can improve the stability of learning and provide meaningful learning curves useful for hyperparameter searches and debugging. However, it is a challenging task to approximate the $K$-Lipschitz constraint which is required by the Wasserstein-1 metric. WGAN-GP \cite{gulrajani2017improved} is proposed by utilizing gradient penalty for restricting $K$-Lipschitz constraint and the objective function is
\begin{eqnarray}\label{equ:30_1}
\begin{array}{l}
 L =  - {E_{x \sim {p_{data}}}}\left[ {D\left( x \right)} \right] + {E_{\tilde x \sim {p_g}}}\left[ {D\left( {\tilde x} \right)} \right] \\ 
 \quad  + \lambda {E_{\hat x \sim {p_{\hat x}}}}\left[ {{{\left( {{{\left\| {{\nabla _{\hat x}}D\left( {\hat x} \right)} \right\|}_2} - 1} \right)}^2}} \right] \\ 
 \end{array}
 \end{eqnarray}
where the first two terms are the objective function of WGAN and ${\hat x}$ is sampled from the distribution ${{p_{\hat x}}}$ which samples uniformly along straight lines between
pairs of points sampled from the real data distribution ${{p_{data}}}$ and the generated distribution ${{p_g}}$. There are some other methods closely related to WGAN-GP such as DRAGAN \cite{kodali2017convergence}. Wu et al. \cite{wu2018wasserstein} propose a novel and relaxed version of Wasserstein-1 metric: Wasserstein divergence (W-div), which does not require the $K$-Lipschitz constraint. Based on W-div, Wu et al. \cite{wu2018wasserstein} introduce a Wasserstein divergence objective for GANs (WGAN-div), which can faithfully approximate W-div by optimization. CramerGAN \cite{bellemare2017cramer} argues that the Wasserstein distance leads to biased gradients, suggesting the Cramér distance between two distributions. Other papers related to WGAN can be found in \cite{petzka2017regularization,juefei2017gang,hsu2017voice,adler2018banach,chen2018deep,athey2019using}.

\subsubsection{Loss Sensitive GAN (LS-GAN)}
Similar to WGAN, LS-GAN \cite{qi2017loss} also has a Lipschitz constraint. It is assumed in LS-GAN that $p_{data}$ lies in a set of Lipschitz densities with a compact support. In LS-GAN , the loss function ${L_\theta }\left( x \right)$ is parameterized with $\theta$ and LS-GAN assumes that a generated sample should have larger loss
than a real one. The loss function can be trained to satisfy the following constraint:
 \begin{eqnarray}\label{equ:LS-GAN1}
 {L_\theta }\left( x \right) \le {L_\theta }\left( {G\left( z \right)} \right) - \Delta \left( {x,G\left( z \right)} \right)
\end{eqnarray}
where $\Delta \left( {x,G\left( z \right)} \right)$ is the margin measuring the difference between generated sample $G(z)$ and real sample $x$. The objective function of LS-GAN is
 \begin{eqnarray}\label{equ:LS-GAN2}
\begin{array}{l}
 \mathop {\min }\limits_D {{\cal L}_D} = {E_{x \sim {p_{data}}\left( x \right)}}\left[ {{L_\theta }\left( x \right)} \right] \\ 
  + \lambda {E_{\scriptstyle x \sim {p_{data}}\left( x \right), \hfill \atop 
  \scriptstyle z \sim {p_z}\left( z \right) \hfill}}{\left[ {\Delta \left( {x,G\left( z \right)} \right) + {L_\theta }\left( x \right) - {L_\theta }\left( {G\left( z \right)} \right)} \right]_ + }, \\ 
 \end{array}
 \end{eqnarray}
 
  \begin{eqnarray}\label{equ:LS-GAN3}
 \mathop {\min }\limits_G {{\cal L}_G} = {E_{z \sim {p_z}\left( z \right)}}\left[ {{L_\theta }\left( {G\left( z \right)} \right)} \right],
 \end{eqnarray}
 where ${\left[ y \right]^ + } = \max (0,y)$, $\lambda$ is the free tuning-parameter, and $\theta$ is the paramter of the discriminator $D$. 

\subsubsection{Summary}
There is a website called ``The GAN Zoo'' (\url{https://github.com/hindupuravinash/the-gan-zoo}) which lists many GANs' variants. Please refer to this website for more details. 

\subsection{GANs Training}\label{subsection:GANsTraining}
Despite the theoretical existence of unique solutions, GANs training is hard and often unstable for several reasons \cite{radford2015unsupervised,salimans2016improved,Arjovsky2017towards}. One difficulty is from the fact that optimal weights for GANs correspond to saddle points, and not minimizers, of the loss function. 

There are many papers on GANs training. Yadav et al. \cite{yadav2017stabilizing} stabilized GANs with prediction methods. By using independent learning rates, \cite{heusel2017gans} proposed a two time-scale update rule (TTUR) for both  discriminator and generator to ensure that the model can converge to a stable local Nash equilibrium. Arjovsky \cite{Arjovsky2017towards} made theoretical steps towards fully understanding the training dynamics of GANs; analyzed why GANs was hard to train; studied and proved rigorously the problems including saturation and instability that occurred when training GANs; examined a practical and theoretically grounded direction to mitigate these problems; and introduced new tools to study them. Liang et al. \cite{liangKevin2018generative} think that GANs training is a continual learning problem \cite{shin2017continual}.

One method to improve GANs training is to assess the empirical ``symptoms'' that might occur in training. These symptoms include: the generative model collapsing to produce very similar samples for diverse inputs \cite{salimans2016improved}; the discriminator loss converging quickly to zero \cite{Arjovsky2017towards}, providing no gradient updates to the generator; difficulties in making the pair of models converge \cite{radford2015unsupervised}.

We will introduce GANs training from three perspectives: objective function, skills, and structure.

\subsubsection{Objective function}
As we can see from Subsection \ref{subsection:GANs}, utilizing the original objective function in equation (\ref{equ:1}) will have the gradient vanishing problem for training $G$ and utilizing the alternative $G$ loss (\ref{equ:12}) in non-saturating game will get the mode collapse problem. These problems are caused by the objective function and cannot be solved by changing the structures of GANs. Re-designing the objective function is a natural solution to mitigate these problems. Based on the theoretical flaws of GANs, many objective function based variants have been proposed to change the objective function of GANs based on theoretical analyses such as least squares generative adversarial networks \cite{mao2017least,mao2019effectiveness}. 

\paragraph{Least squares generative adversarial networks (LSGANs) }~{}\newline
LSGANs \cite{mao2017least,mao2019effectiveness} are proposed to overcome the vanishing gradient problem in the original GANs. This work shows that
the decision boundary for $D$ of original GAN penalizes very small error to update $G$ for those generated samples which are far from the decision boundary. LSGANs adopt the least squares loss rather than the cross-entropy loss in the original GANs. Suppose that the $a$-$b$ coding is used for the LSGANs' discriminator \cite{mao2017least}, where $a$ and $b$ are the labels for generated sample and real sample, respectively. The LSGANs' discriminator loss ${V_{LSGAN}}\left( D \right)$ and generator loss ${V_{LSGAN}}\left( G \right)$ are defined as:
\begin{eqnarray}\label{equ:31}
\begin{array}{l}
 \mathop {\min }\limits_D \;{V_{LSGAN}}\left( D \right) = {E_{x \sim {p_{data}}\left( x \right)}}\left[ {{{\left( {D\left( x \right) - b} \right)}^2}} \right] \\ 
 \quad \quad \quad \quad \quad \, + {E_{z \sim {p_z}\left( z \right)}}\left[ {{{\left( {D\left( {G\left( z \right)} \right) - a} \right)}^2}} \right], \\ 
 \end{array}
 \end{eqnarray}
 
\begin{eqnarray}\label{equ:32}
\mathop {\min }\limits_G \;{V_{LSGAN}}\left( G \right) = {E_{z \sim {p_z}\left( z \right)}}\left[ {{{\left( {D\left( {G\left( z \right)} \right) - c} \right)}^2}} \right],
\end{eqnarray}
where $c$ is the value that  $G$ hopes for $D$ to believe for generated samples. The reference \cite{mao2017least} shows that there are two advantages of LSGANs in comparison with the original GANs:
\begin{itemize}
    \item The new decision boundary produced by $D$ penalizes large error to those generated samples which are far from the decision boundary, which makes those ``low quality'' generated samples move toward the decision boundary. This is good for generating higher quality samples.
    \item Penalizing the generated samples far  from the decision boundary can supply more gradient when updating the $G$, which overcomes the vanishing gradient problems in the original GANs.
\end{itemize}

\paragraph{Hinge loss based GAN}~{}\newline
Hinge loss based GAN is proposed and used in \cite{miyato2018spectral,lim2017geometric,tran2017deep} and its objective function is $V\left( {D,G} \right)$:
\begin{eqnarray}\label{equ:33}
\begin{array}{l}
 {V_D}\left( {\hat G,D} \right) = {E_{x \sim {p_{data}}\left( x \right)}}\left[ {\min (0, - 1 + D\left( x \right))} \right] \\ 
 \quad \quad  + {E_{z \sim {p_z}\left( z \right)}}\left[ {\min (0, - 1 - D\left( {\hat G(z)} \right))} \right]. \\ 
 \end{array}
 \end{eqnarray}
 
\begin{eqnarray}\label{equ:34}
{V_D}\left( {G,\hat D} \right) =  - {E_{z \sim {p_z}\left( z \right)}}\left[ {\hat D\left( {G(z)} \right)} \right].
\end{eqnarray}
The softmax cross-entropy loss \cite{lin2017softmax} is also used in GANs. 

 \paragraph{Energy-based generative adversarial network (EBGAN)}~{}\newline
EBGAN's discriminator is seen as an energy function, giving high energy to the fake (``generated'') samples and lower energy to the real samples. As for the energy function, please refer to \cite{lecun2006tutorial} for the corresponding tutorial. Given a positive margin $m$ , the loss functions for EBGAN can be defined as follows:
\begin{eqnarray}\label{equ:35}
{{\cal L}_D}(x,z) = D(x) + {\left[ {m - D(G(z))} \right]^ + },
\end{eqnarray}

\begin{eqnarray}\label{equ:36}
{{\cal L}_G}(z) = D(G(z)),
\end{eqnarray}
where ${\left[ y \right]^ + } = \max (0,y)$ is the rectified linear unit (ReLU) function. Note that in the original GANs, the discriminator $D$ give high score to real samples and low score to the generated (``fake'') samples. However, the discriminator in EBGAN attributes low energy (score) to the real samples and higher energy to the generated ones. EBGAN has more stable behavior than original GANs during training.

 \paragraph{Boundary equilibrium generative adversarial networks (BEGAN)}~{}\newline
Similar to EBGAN \cite{zhao2017energy}, dual-agent GAN (DA-GAN) \cite{zhao2017dual,zhao20183d}, and margin adaptation for GANs (MAGANs) \cite{wang2017magan}, BEGAN also uses an auto-encoder as the discriminator. Using proportional control theory, BEGAN proposes a novel equilibrium method to balance generator and discriminator in training, which is fast, stable, and robust to parameter changes.

\paragraph{Mode regularized generative adversarial networks (MDGAN) }\label{paragraph:mode}~{}\newline
Che et al. \cite{che2017mode} argue that GANs' unstable training and model collapse is due to the very special functional shape of the trained discriminators in high dimensional spaces, which can make training stuck or push probability mass in the wrong direction, towards that of higher concentration than that of the real data distribution. Che et al. \cite{che2017mode} introduce several methods of regularizing the objective, which can stabilize the training of GAN models. The key idea of MDGAN is utilizing an encoder $E\left( x \right):x \to z$ to produce the latent variable $z$ for the generator $G$ rather than utilizing noise. This procedure has two advantages: 
\begin{itemize}
\item Encoder guarantees the correspondence between $z$ ($E(x)$) and $x$, which makes $G$ capable of covering diverse modes in the data space. Therefore, it prevents the mode collapse problem.
\item Because the reconstruction of encoder can add more information to the generator $G$, it is not easy for the discriminator $D$ to distinguish between real samples and generated ones. 
\end{itemize}
The loss function for the generator and the encoder of MDGAN is
\begin{eqnarray}\label{equ:MDGAN1}
\begin{array}{l}
 {{\cal L}_G} =  - {E_{z \sim {p_z}\left( z \right)}}\left[ {\log \left( {D\left( {G\left( z \right)} \right)} \right)} \right] \\ 
  + {E_{x \sim {p_{data}}\left( x \right)}}\left[ \begin{array}{l}
 {\lambda _1}d\left( {x,G \circ E\left( x \right)} \right) \\ 
  + {\lambda _2}\log D\left( {G \circ E\left( x \right)} \right) \\ 
 \end{array} \right], \\ 
 \end{array}
\end{eqnarray}

\begin{eqnarray}\label{equ:MDGAN2}
{{\cal L}_E} = {E_{x \sim {p_{data}}\left( x \right)}}\left[ \begin{array}{l}
 {\lambda _1}d\left( {x,G \circ E\left( x \right)} \right) \\ 
  + {\lambda _2}\log D\left( {G \circ E\left( x \right)} \right) \\ 
 \end{array} \right],
\end{eqnarray}
where both $\lambda _1$ and $\lambda _2$ are free tuning parameters, $d$ is the distance metric such as Euclidean distance, and $G \circ E\left( x \right) = G\left( {E\left( x \right)} \right)$.

\paragraph{Unrolled GAN}~{}\newline
Metz et al. \cite{metz2016unrolled} introduce a technique to stabilize GANs by defining the generator objective with regard to an unrolled optimization of the discriminator. This allows training to be adjusted between utilizing the current value of the discriminator, which is usually unstable and leads to poor solutions, and utilizing the optimal discriminator solution in the generator's objective, which is perfect but infeasible in real applications. Let $f\left( {{\theta _G},{\theta _D}} \right)$ denote the objective function of the original GANs. 

A local optimal solution of the discriminator parameters $\theta _D^*$ can be expressed as the fixed point of an iterative optimization procedure,
\begin{eqnarray}\label{equ:Unrolled1}
\theta _D^0 = {\theta _D},
\end{eqnarray}

\begin{eqnarray}\label{equ:Unrolled2}
\theta _D^{k + 1} = \theta _D^k + {\eta ^k}\frac{{df\left( {{\theta _G},\theta _D^k} \right)}}{{d\theta _D^k}},
\end{eqnarray}

\begin{eqnarray}\label{equ:Unrolled3}
\theta _D^*\left( {{\theta _G}} \right) = \mathop {\lim }\limits_{k \to \infty } \theta _D^k,
\end{eqnarray}
where ${\eta ^k}$ is the learning rate. By unrolling for $K$ steps, a surrogate objective for the update of the generator is created
\begin{eqnarray}\label{equ:Unrolled4}
{f_K}\left( {{\theta _G},{\theta _D}} \right) = f\left( {{\theta _G},\theta _D^K\left( {{\theta _G},{\theta _D}} \right)} \right).
\end{eqnarray}
When $K = 0$, this objective is the same as the standard GAN objective. When $K \to \infty $, this objective is the true generator objective function $f\left( {{\theta _G},\theta _D^*\left( G \right)} \right)$. By adjusting the number of
unrolling steps $K$, we are capable of interpolating between standard GAN training dynamics with their related pathologies, and more expensive gradient descent on the true generator loss. The generator and discriminator parameter updates of unrolled GAN using this surrogate loss are
\begin{eqnarray}\label{equ:Unrolled5}
{\theta _G} = {\theta _G} - \eta \frac{{d{f_K}\left( {{\theta _G},{\theta _D}} \right)}}{{d{\theta _G}}},
\end{eqnarray}
\begin{eqnarray}\label{equ:Unrolled6}
{\theta _D} = {\theta _D} + \eta \frac{{df\left( {{\theta _G},{\theta _D}} \right)}}{{d{\theta _D}}}.
\end{eqnarray}
Metz et al. \cite{metz2016unrolled} show how this method solves mode collapse, stabilizes training of GANs, and increases diversity and coverage of the generated distribution by the generator.

\paragraph{Spectrally normalized GANs (SN-GANs)}~{}\newline
SN-GANs \cite{miyato2018spectral} propose a novel weight normalization method named spectral normalization to make the training of the discriminator stable. This new normalization technique is computationally efficient and easy to be integrated into existing methods. The spectral normalization \cite{miyato2018spectral} uses a simple method to make the weight matrix $W$ satisfy the Lipschitz constraint $\sigma \left( W \right) = 1$:
\begin{eqnarray}\label{equ:SN-GANs1}
{\bar W_{SN}}\left( W \right): = W/\sigma \left( W \right),
\end{eqnarray}
where $W$ is the weight matrix of each layer in $D$ and $\sigma \left( W \right)$ is the spectral norm of $W$. It is shown that \cite{miyato2018spectral} SN-GANs can generate images of equal or better quality in comparison with the previous training stabilization methods. In theory, spectral normalization is capable of being applied to all GANs variants. Both BigGANs \cite{brock2018large} and SAGAN \cite{zhang2018self} use the spectral normalization and have good performances on the Imagenet.

\paragraph{Relativistic GANs (RGANs)}~{}\newline
In the original GANs, the discriminator can be defined, according to the non-transformed layer $C(x)$, as $D(x) = sigmoid(C(x))$. A simple way to make discriminator relativistic (i.e., making the output of $D$ depend on both real and generated samples) \cite{jolicoeur2018relativistic} is to sample from real/generated data pairs $\tilde x = \left( {{x_r},{x_g}} \right)$ and define it as \begin{eqnarray}\label{equ:RGANs0}
D\left( {\tilde x} \right) = sigmoid\left( {C\left( {{x_r}} \right) - C\left( {{x_g}} \right)} \right).
\end{eqnarray}
This modification can be understood in the following way \cite{jolicoeur2018relativistic}:
$D$ estimates the probability that the given real sample is more realistic than a randomly sampled generated sample. Similarly, ${D_{rev}}\left( {\tilde x} \right) = sigmoid\left( {C\left( {{x_g}} \right) - C\left( {{x_r}} \right)} \right)$ can be defined as the probability that the given generated sample is more realistic than a randomly sampled real sample. The discriminator and generator loss functions of the Relativistic Standard GAN
(RSGAN) is:
\begin{eqnarray}\label{equ:RGANs1}
L_D^{RSGAN} =  - {E_{({x_r},{x_g}\;)}}\left[ {\log \left( {sigmoid\left( {C\left( {{x_r}} \right) - C\left( {{x_g}} \right)} \right)} \right)} \right],
\end{eqnarray}
\begin{eqnarray}\label{equ:RGANs2}
L_G^{RSGAN} =  - {E_{({x_r},{x_g}\;)}}\left[ {\log \left( {sigmoid\left( {C\left( {{x_g}} \right) - C\left( {{x_r}} \right)} \right)} \right)} \right].
 \end{eqnarray}
 Most GANs can be parametrized:
 \begin{eqnarray}\label{equ:RGANs3}
 L_D^{GAN} = {E_{{x_r}}}\left[ {{f_1}\left( {C\left( {{x_r}} \right)} \right)} \right] + {E_{{x_g}}}\left[ {{f_2}\left( {C\left( {{x_g}} \right)} \right)} \right],
 \end{eqnarray}
  \begin{eqnarray}\label{equ:RGANs4}
 L_G^{GAN} = {E_{{x_r}}}\left[ {{g_1}\left( {C\left( {{x_r}} \right)} \right)} \right] + {E_{{x_g}}}\left[ {{g_2}\left( {C\left( {{x_g}} \right)} \right)} \right],
 \end{eqnarray}
where $f_1$, $f_2$, $g_1$, $g_2$ are scalar-to-scalar functions. If we adopt a relativistic discriminator, the loss functions of these GANs become:
\begin{eqnarray}\label{equ:RGANs5}
\begin{array}{l}
 L_D^{RGAN} = {E_{({x_r},{x_g}\;)}}\left[ {{f_1}\left( {C\left( {{x_r}} \right) - C\left( {{x_g}} \right)} \right)} \right] \\ 
 \quad \quad  + {E_{({x_r},{x_g}\;)}}\left[ {{f_2}\left( {C\left( {{x_g}} \right) - C\left( {{x_r}} \right)} \right)} \right] \\ 
 \end{array},
\end{eqnarray}
\begin{eqnarray}\label{equ:RGANs6}
\begin{array}{l}
 L_G^{RGAN} = {E_{({x_r},{x_g}\;)}}\left[ {{g_1}\left( {C\left( {{x_r}} \right) - C\left( {{x_g}} \right)} \right)} \right] \\ 
 \quad \quad  + {E_{({x_r},{x_g}\;)}}\left[ {{g_2}\left( {C\left( {{x_g}} \right) - C\left( {{x_r}} \right)} \right)} \right]. \\ 
 \end{array}
\end{eqnarray}

\subsubsection{Skills}
NIPS 2016 held a workshop on adversarial training, with an invited talk by Soumith Chintala named "How to train a GAN." This talk has an assorted tips and tricks. For example, this talk suggests that if you have labels, training the discriminator to also classify the examples: AC-GAN \cite{odena2017conditional}. Refer to the GitHub repository associated with Soumith's talk: \url{https://github.com/soumith/ganhacks} for more advice. 

Salimans et al. \cite{salimans2016improved} proposed very useful and improved techniques for training GANs (ImprovedGANs), such as feature matching, minibatch discrimination, historical averaging, one-sided label smoothing, and virtual batch normalization. 

\subsubsection{Structure}
The original GANs utilized multi-layer perceptron (MLP). Specific type of structure may be good for specific applications e.g., recurrent neural network (RNN) for time series data and convolutional neural network (CNN) for images.

\paragraph{The original GANs}~{}\newline
The original GANs used MLP as the generator $G$ and discriminator $D$. MLP can be only used for small datasets such as CIFAR-10 \cite{krizhevsky2009learning}, MNIST \cite{lecun1998gradient}, and the Toronto Face Database (TFD) \cite{susskind2010toronto}. However, MLP does not have good generalization on more
complex images \cite{wang2019generative}.

\paragraph{Laplacian generative adversarial networks (LAPGAN) and SinGAN}~{}\newline
LAPGAN \cite{denton2015deep} is proposed for producing higher resolution images in comparison with the original GANs. LAPGAN uses a cascade of CNN
within a Laplacian pyramid framework \cite{burt1983laplacian} to generate high quality images.

SinGAN \cite{shaham2019singan} learns a generative model from a single natural image. SinGAN has a pyramid of fully convolutional GANs, each of which learns the patch distribution at a different scale of the image. Similar to SinGAN, InGAN \cite{shocher2019ingan} also learns a generative model from a single natural image.

\paragraph{Deep convolutional generative adversarial networks (DCGANs)}~{}\newline
In original GANs, $G$ and $D$ are defined by MLP. Because CNN are better at images than MLP, $G$ and $D$ are defined by deep convolutional neural networks (DCNNs) in DCGANs \cite{radford2015unsupervised}, which have better performance. Most current GANs are at least loosely based on the DCGANs architecture \cite{radford2015unsupervised}. Three key features of the DCGANs architecture are listed as follows:
\begin{itemize}
\item First, the overall architecture is mostly based on the all-convolutional
net \cite{springenberg2014striving}. This architecture has neither pooling
nor ``unpooling'' layers. When $G$ needs to increase the spatial dimensionality of the representation, it uses transposed convolution (deconvolution) with a stride greater than 1.
\item Second, utilize batch normalization for most layers of both $G$ and $D$. The last layer of $G$ and first layer of $D$ are not batch normalized, in order that the neural network can learn the correct mean and scale of the data distribution.
\item Third, utilize the Adam optimizer instead of SGD with momentum.
\end{itemize}

\paragraph{Progressive GAN}~{}\newline
In Progressive GAN (PGGAN) \cite{karras2017progressive}, a new training methodology for GAN is proposed. The structure of Progressive GAN is based on progressive
neural networks that is first proposed in \cite{rusu2016progressive}. The key idea of Progressive GAN is to grow both the generator and discriminator progressively: starting from a low resolution, adding new layers that model increasingly fine details as training progresses. 

\paragraph{Self-Attention Generative Adversarial Network (SAGAN)}~{}\newline
SAGAN \cite{zhang2018self} is proposed to allow attention-driven, long-range dependency modeling for image generation tasks. Spectral normalization technique has only been applied to the discriminator \cite{miyato2018spectral}. SAGAN uses spectral normalization for both generator and discriminator and it is found that this improves training dynamics. Furthermore, it is confirmed that the two time-scale update rule (TTUR) \cite{heusel2017gans} is effective in SAGAN.

Note that AttnGAN \cite{xu2018attngan} utilizes attention over word embeddings within an input sequence rather than self-attention over internal model states.

\paragraph{BigGANs and StyleGAN}~{}\newline
Both BigGANs \cite{brock2018large} and StyleGAN \cite{karras2019style,karras2019analyzing} made great advances in the quality of GANs. 

BigGANs \cite{brock2018large} is a large scale TPU implementation of GANs, which is pretty similar to SAGAN but scaled up greatly. BigGANs successfully generates images with quite high resolution up to 512 by 512 pixels. If you do not have enough data, it can be a challenging task to replicate the BigGANs results from scratch. Lucic et al. \cite{luvcic2019high} propose to train BigGANs quality model with fewer labels. BigBiGAN \cite{Jeff2019Large}, based on BigGANs, extends it to representation learning by adding an encoder and modifying the discriminator. BigBiGAN achieve the state of the art in both unsupervised representation learning on ImageNet and unconditional image generation.

In the original GANs \cite{goodfellow2014generative}, $G$ and $D$ are defined by MLP. Karras et al. \cite{karras2019style} proposed a StyleGAN architecture for GANs, which wins the CVPR 2019 best paper honorable mention. StyleGAN's generator is a really high-quality generator for other generation tasks like generating faces. It is particular exciting because it allows to separate different factors such as hair, age and sex that are involved in controlling the appearance of the final example and we can then control them separately from each other. StyleGAN \cite{karras2019style} has also been used in such as generating high-resolution fashion model images wearing custom outfits \cite{yildirim2019generating}.

\paragraph{Hybrids of autoencoders and GANs}~{}\label{paragraph:autoencoders}\newline
An autoencoder is a type of neural networks used for learning efficient data codings in an unsupervised way. The autoencoder has an encoder and a decoder. The encoder aims to learn a representation (encoding) for a set of data, $z=E(x)$, typically for dimensionality reduction. The decoder aims to reconstruct the data $\hat x = g\left( z \right)$. That is to say, the decoder tries to generate from the reduced encoding a representation as close as possible to its original input $x$.

\textbf{GANs with an autoencoder}: Adversarial autoencoder (AAE) \cite{makhzani2015adversarial} is a probabilistic autoencoder based on GANs. Adversarial variational Bayes (AVB) \cite{mescheder2017adversarial,yuvaegan} is proposed to unify variational autoencoders (VAEs) and GANs. Sun et al. \cite{liu2017unsupervised} proposed a UNsupervised Image-to-image Translation (UNIT) framework that are based on GANs and VAEs. Hu et al. \cite{hu2017unifying} aimed to establish formal connections between GANs and VAEs through a new formulation of them. By combining a VAE with a GAN, Larsen et al. \cite{larsen2015autoencoding} utilize learned feature representations in the GAN discriminator as basis for the VAE reconstruction. Therefore, element-wise errors is replaced with feature-wise errors to better capture the data distribution while offering invariance towards such as translation. Rosca et al. \cite{rosca2017variational} proposed variational approaches for auto-encoding GANs. By adopting an encoder-decoder architecture for the generator, disentangled representation GAN (DR-GAN) \cite{tran2018representation} addresses pose-invariant face recognition, which is a hard problem due to the drastic changes in an image for each diverse pose.

\begin{figure}
\centering
\subfloat[BiGAN/ALI] {\scalebox{0.42}{\includegraphics{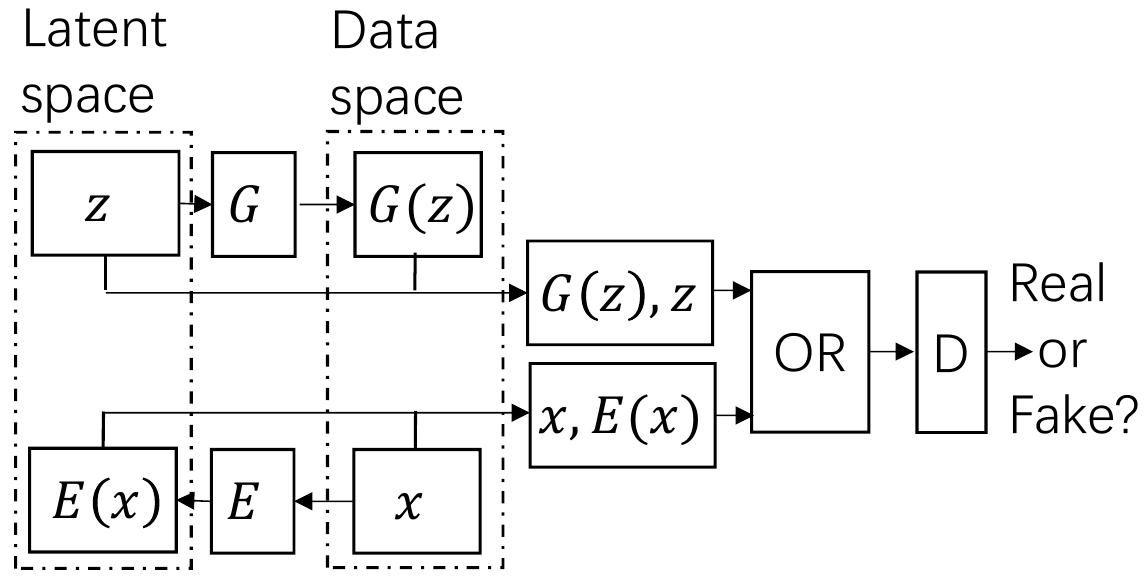}}}
\subfloat[AGE] {\scalebox{0.38}{\includegraphics{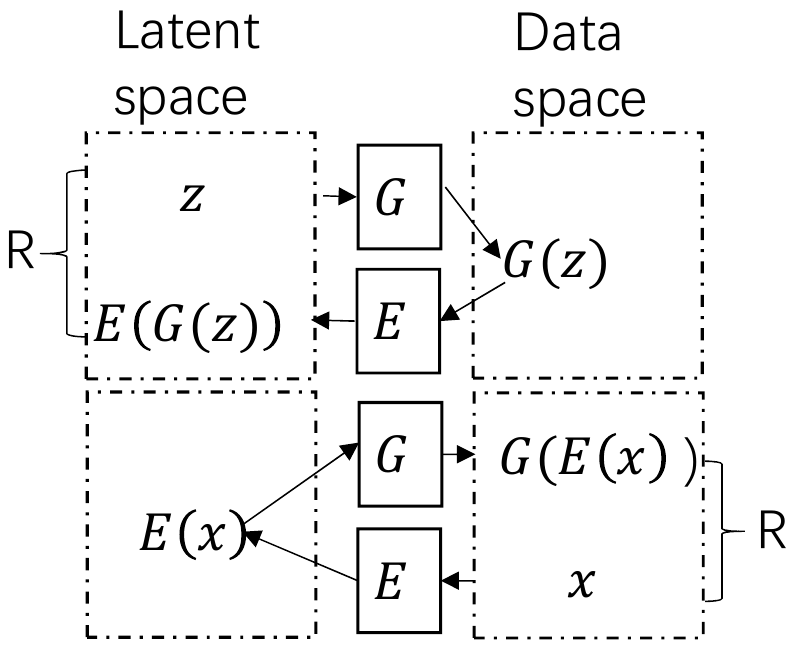}}}
\caption{The structures of (a) BiGAN and ALI and (b) AGE.}
\label{Fig:3_1}
\end{figure}

\textbf{GANs with an encoder}: References \cite{donahue2016adversarial,ulyanov2018takes} only add an encoder to GANs. The original GANs \cite{goodfellow2014generative}
can not learn the inverse mapping - projecting data back into the latent space. To solve this problem, Donahue et al. \cite{donahue2016adversarial} proposed  Bidirectional GANs (BiGANs), which can learn this inverse mapping through the encoder, and show that the resulting learned feature representation is useful. Similarly, Dumoulin et al. \cite{dumoulin2016adversarially} proposed the adversarially learned inference (ALI) model, which also utilizes the encoder to learn the latent feature distribution. The structure of BiGAN and ALI is shown in Fig. \ref{Fig:3_1}(a). Besides the discriminator and generator, BiGAN also has an encoder, which is used for mapping the data back to the latent space. The input of discriminator is a pair of data composed of data and its corresponding latent code. For real data $x$, the pair are $x, E(x)$ where $E(x)$ is obtained from the encoder $E$. For the generated data $G(z)$, this pair is $G(z), z$ where $z$ is the noise vector generating the data $G(z)$ through the generator $G$. Similar to (\ref{equ:1}), the objective function of BiGAN is
\begin{eqnarray}\label{equ:BiGANs1}
\begin{array}{l}
 \mathop {\min }\limits_{G,E} {\rm{ }}\mathop {\max }\limits_D \;V\left( {D,E,G} \right) = {E_{x \sim {p_{data}}\left( x \right)}}\left[ {\log D\left( {x,E(x)} \right)} \right] \\ 
 \quad  + {E_{z \sim {p_z}\left( z \right)}}\left[ {\log \left( {1 - D\left( {G\left( z \right),z} \right)} \right)} \right]. \\ 
 \end{array}
\end{eqnarray}
The generator of \cite{donahue2016adversarial,ulyanov2018takes} can be seen the decoder since the generator map the vectors from latent space to data space, which performs the same function as the decoder.

Similar to utilizing an encoding process to
model the distribution of latent samples, Gurumurthy et al. \cite{gurumurthy2017deligan} model the latent space as a mixture of Gaussians and learn the mixture components that maximize the likelihood of generated samples under the data generating distribution.
    
In an encoding-decoding model, the output (also known as a reconstruction), ought to be similar to the input in the ideal case. Generally,
the fidelity of reconstructed samples synthesized utilizing a BiGAN/ALI is poor. With an additional adversarial cost on the distribution
of data samples and their reconstructions \cite{li2017alice}, the fidelity of samples may be improved. Other related methods include such as variational discriminator bottleneck (VDB) \cite{peng2018variational} and MDGAN (detailed in Paragraph \ref{paragraph:mode}).

\textbf{Combination of a generator and an encoder}: Different from previous hybrids of autoencoders and GANs, Adversarial Generator-Encoder (AGE) Network \cite{ulyanov2018takes} is set up directly between the generator and the encoder, and no external mappings are trained in the
learning process. The structure of AGE is shown in Fig. \ref{Fig:3_1}(b) which $R$ is the reconstruction loss function. In AGE, there are two reconstruction losses: the latent variable $z$ and $E(G(z))$, the data $x$ and $G(E(x))$. AGE is similar to CycleGAN. However, there are two differences between them:
\begin{itemize}
\item CycleGAN \cite{zhu2017unpaired} is used for two modalities of the image such as grayscale and color. AGE acts between latent space and true data space.
\item There is a discriminator for each modality of CycleGAN and there is no discriminator in AGE.
\end{itemize}

\paragraph{Multi-discriminator learning}~{}\newline
GANs have a discriminator together with a generator. Different from GANs, dual discriminator GAN (D2GAN) \cite{nguyen2017dual} has \textbf{a generator and two binary discriminators}. D2GAN is analogous to a minimax game, wherein one discriminator gives high scores for samples from generated distribution whilst the other discriminator, conversely, favoring data from the true distribution, and the generator generates data to fool both discriminators. The reference \cite{nguyen2017dual} develops theoretical analysis to show that, given the optimal discriminators, optimizing the generator of D2GAN is minimizing both KL and reverse KL divergences between true distribution and the generated distribution, thus effectively overcoming the mode collapsing problem. Generative multi-adversarial network (GMAN) \cite{durugkar2016generative} further extends GANs to a generator and multiple discriminators. Albuquerque et al. \cite{albuquerque2019multi} performed multi-objective training of GANs with multiple discriminators. 

\paragraph{Multi-generator learning}~{}\newline
Multi-generator GAN (MGAN) \cite{hoang2017multi} is proposed to train the GANs with a mixture of generators to avoid the mode collapsing problem. More specially, MGAN has \textbf{one binary discriminator, $K$ generators, and a multi-class classifier}. The distinguishing feature of MGAN is that the generated samples are produced from multiple generators, and then only one of them will be randomly selected as the final output similar to the mechanism of a probabilistic mixture model. The classifier shows which generator a generated sample is from. 

The most closely related to MGAN is multi-agent diverse GANs (MAD-GAN) \cite{ghosh2018multi}. The difference between MGAN and MAD-GAN can be found in \cite{hoang2017multi}.
SentiGAN \cite{wang2018sentigan} uses a mixture of generators and a multi-class discriminator to generate sentimental texts.

\begin{figure}
\centering
\subfloat[GANs]{\scalebox{0.48}{\includegraphics{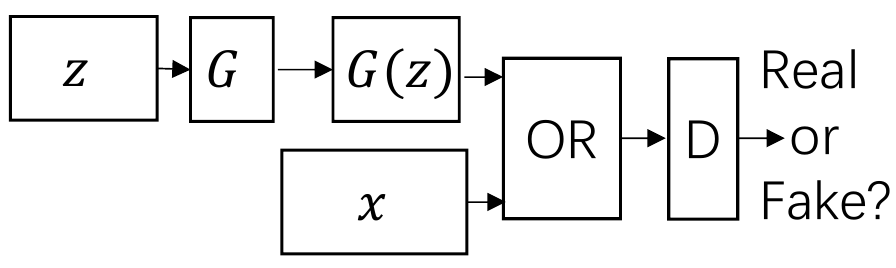}}}\hspace{30pt}
\subfloat[D2GAN] {\scalebox{0.48}{\includegraphics{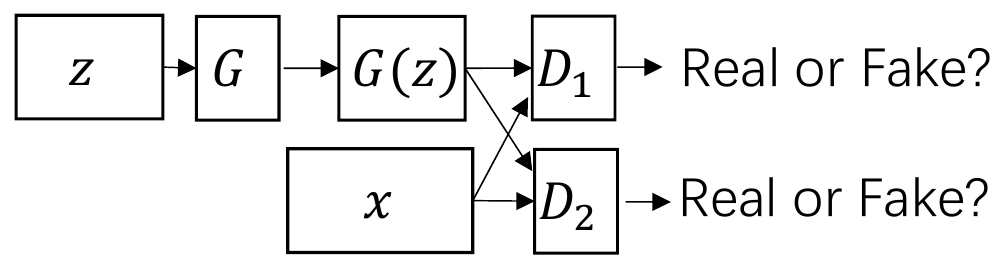}}}\\
\subfloat[MGAN] {\scalebox{0.46}{\includegraphics{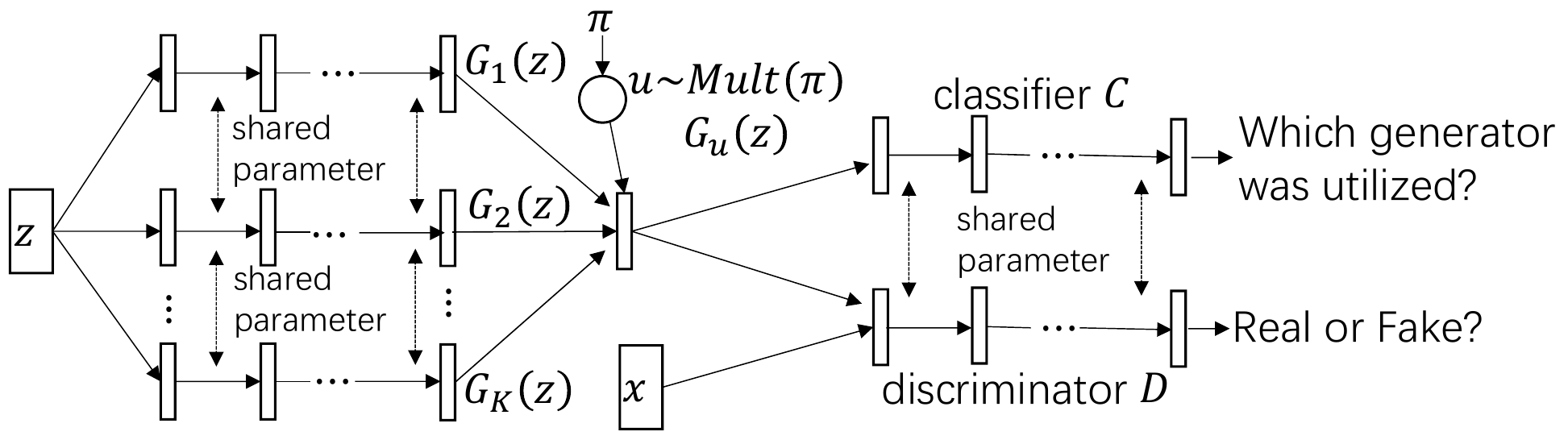}}}\\
\subfloat[CoGAN] {\scalebox{0.48}{\includegraphics{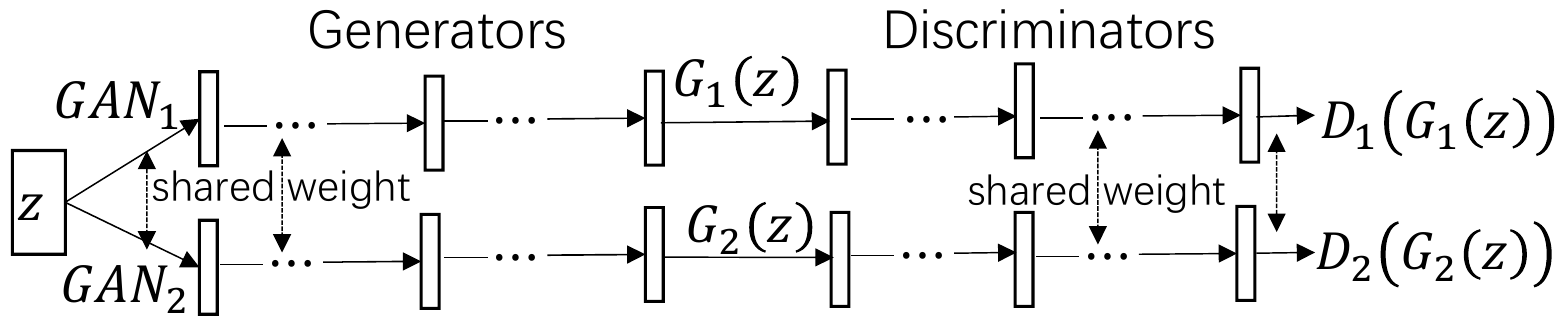}}}
\caption{The structures of GANs \cite{goodfellow2014generative}, D2GAN\cite{nguyen2017dual}, MGAN \cite{hoang2017multi}, and CoGAN \cite{liu2016coupled}.}
\label{Fig:4}
\end{figure}

\paragraph{Multi-GAN learning}~{}\newline
Coupled GAN (CoGAN) \cite{liu2016coupled} is proposed for learning a joint distribution of two-domain images. CoGAN is composed of a pair of GANs - GAN1 and GAN2, each of which synthesizes images in one domain. Two GANs respectively considering structure and style are proposed in \cite{wang2016generative} based on cGANs. Causal fairness-aware GANs (CFGAN) \cite{xu2019CFGAN} used two generators and two discriminators for generating fair data. The structures of GANs, D2GAN, MGAN, and CoGAN are shown in Fig. \ref{Fig:4}.

\begin{figure*}
\centering
\scalebox{0.68}{\includegraphics{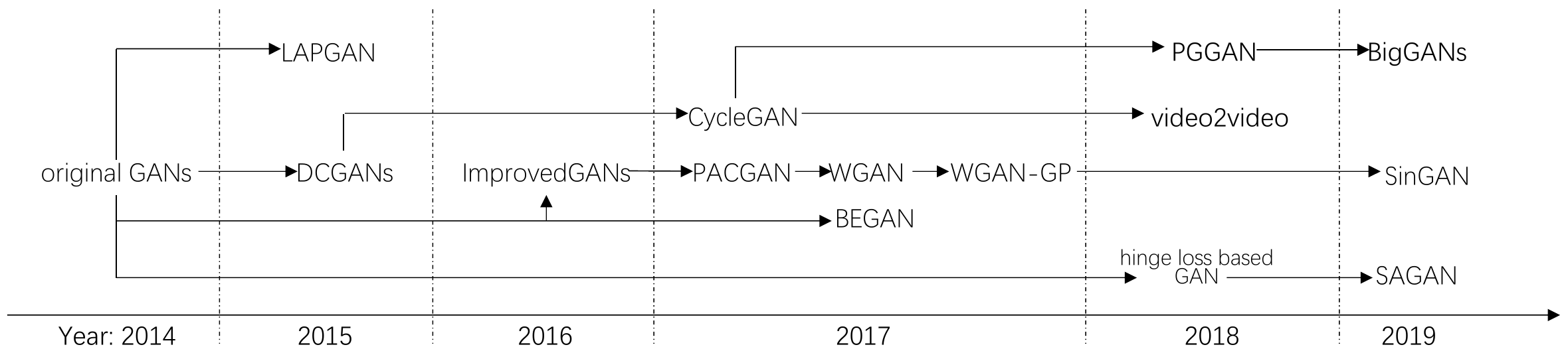}}\\
\caption{A road map of GANs. Milestone variants are shown in this figure.}
\label{Fig:4_1}
\end{figure*}

\paragraph{Summary}~{}\newline
There are many GANs' variants and milestone ones are shown in Fig. \ref{Fig:4_1}. Due to space limitation, only limited number of variants are shown in Fig. \ref{Fig:4_1}. 

GANs' objective function based variants can be generalized to structure variants. Compared with other objective function based variants, both SN-GANs and RGANs show the stronger generalization ability. These two objective function based variants can be generalized to the other
objective function based variants. Spectral normalization is capable of being generalized to any type of GANs' variants while RGANs is able to be generalized to any IPM-based GANs.

\subsection{Evaluation metrics for GANs}\label{subsection:How}
In this subsection, we show evaluation metrics \cite{wang2018use,wang2019neuroscore} that are used for GANs.

\subsubsection{Inception Score (IS)}
Inception score (IS) is proposed in \cite{salimans2016improved}, which uses the Inception model \cite{szegedy2016rethinking} for every generated image to get the conditional label distribution $p\left( {y|x} \right)$. Images that have meaningful objects ought to have a conditional label distribution $p\left( {y|x} \right)$ with low entropy. Furthermore, the model is expected to produce diverse images. Therefore, the marginal $\int {p\left( {y|x = G\left( z \right)} \right)} dz$ ought to have high entropy. In combination of these two requirements, the IS is: 
\begin{eqnarray}\label{equ:IS1}
\exp ({E_x}KL\left( {p\left( {y|x} \right)||p\left( y \right)} \right)),
\end{eqnarray}
where exponentiating results is for easy comparison of the values.

A higher IS indicates that the generative model can produce high quality samples and the generated samples are also diverse. However, the IS also has disadvantages. If the generative model falls into mode collapse, the IS might be still good while the real case is pretty bad. To address this issue, an independent Wasserstein critic \cite{danihelka2017comparison} is proposed to be trained independently for the validation dataset to measure mode collapse and overfitting.

\subsubsection{Mode score (MS)}
The mode score (MS) \cite{xu2018empirical,che2017mode} is an improved version of the IS. Different from IS, MS can measure the dissimilarity between the real distribution and generated distribution.

\subsubsection{Fr\'echet Inception Distance (FID)}
FID was also proposed \cite{heusel2017gans} to evaluate
GANs. For a suitable feature function $\phi$
(the default one is the Inception network's convolutional feature), FID models $\phi \left( {{p_{data}}} \right)$ and $\phi \left( {{p_g}} \right)$ as Gaussian random variables with empirical means ${\mu _r},\;{\mu _g}$ and
empirical covariance ${C_r},\;{C_g}$ and computes
\begin{eqnarray}\label{equ:FID1}
\begin{array}{l}
 FID({p_{data}},{p_g}) = \left\| {{\mu _r} - \;{\mu _g}} \right\| \\ 
 \quad \; + tr\left( {{C_r} + \;{C_g} - 2{{\left( {{C_r}{C_g}} \right)}^{1/2}}} \right), \\ 
 \end{array}
\end{eqnarray}
which is the Fr\'echet distance (also called the Wasserstein-2 distance) between the two Gaussian
distributions. However, the IS and FID cannot well handle the overfitting problem. To mitigate this problem, the kernel inception distance (KID) was proposed in \cite{binkowski2018demystifying}.

\subsubsection{Multi-scale structural similarity (MS-SSIM)}
Structural similarity (SSIM) \cite{wang2004image} is proposed to measure the similarity between two images. Different from single scale SSIM measure, the MS-SSIM \cite{wang2003multiscale} is proposed for multi-scale image quality assessment. It quantitatively evaluates image similarity by attempting to predict human perceptual similarity judgment. The range of MS-SSIM
values is between 0.0 and 1.0 and lower MS-SSIM values means perceptually more dissimilar images. References \cite{odena2017conditional,fedus2017many} used MS-SSIM to measure the diversity of fake data.
Reference \cite{kurach2019large} suggested that MS-SSIM should only be taken into account together with the FID and IS metrics for testing
sample diversity.

\subsubsection{Summary}
How to select a good evaluation metric for GANs is still a hard problem \cite{theis2015note}. Xu et al. \cite{xu2018empirical} proposed an empirical study on evaluation metrics of GANs. Karol Kurach \cite{kurach2019large} conducted a large-scale study on regularization and normalization in GANs. There are some other comparative study of GANs such as \cite{lucic2018gans}. Reference \cite{borji2019pros} presented several measures as meta-measures to guide researchers to select quantitative evaluation metrics. An appropriate evaluation metric ought to differentiate true samples from fake ones, verify mode drop, mode collapse, and detect overfitting. It is hoped that there will be better methods to evaluate the quality of the GANs model in the future.

\subsection{Task driven GANs}
The focus of this paper is on GANs. There are closely associated fields for specific tasks with an enormous volume of literature.
\subsubsection{Semi-Supervised Learning}
A research field where GANs are very successful is the application of generative models to semi-supervised learning \cite{denton2016semi,ding2018semi}, as proposed but not shown in the first GANs paper \cite{goodfellow2014generative}. 

GANs have been successfully used for semi-supervised learning at least since CatGANs \cite{springenberg2015unsupervised}. Feature matching GANs \cite{salimans2016improved} got good performance with a small number of labels on datasets such as MNIST, SVHN, and CIFAR-10.

Odena \cite{odena2016semi} extends GANs to the semi-supervised learning by forcing the discriminator network to output class labels. Generally speaking, when we train GANs, we actually do not use the discriminator in the end. The discriminator is only used to guide the learning process, but the discriminator is not used to generate the data after we have trained the generator. We only use the generator to generate the data and abandon discriminator at last. For traditional GANs, the discriminator is a two-class classifier, which outputs category one for real data and category two for generated data. In semi-supervised learning, the discriminator is upgraded to be a multi-class classifier. At the end of the training, the classifier is the thing that we are interested in. For semi-supervised learning, if we want to learn an $N$ class classifier, we make GANs with a discriminator which can predict $N$+1 classes the input is from, where an extra class corresponds to the outputs of $G$. Therefore, suppose that we want to learn to classify two classes, apples and oranges. We can make a classifier which has three different labels: one - the class of real apples, two - the class of real oranges, and three - the class of generated data. The system learns on three kinds of data: real labeled data, unlabeled real data, and fake data. 

\textbf{Real labeled data}: We can tell the discriminator to maximize the probability of the correct class. For example, if we have an apple photo and it is labeled as an apple, we can maximize the probability of the apple class in this discriminator. 

\textbf{Unlabeled real data}: Suppose we have a photo, we do not know whether it is an apple or an orange but we know that it is a real photo. In this situation, we train the discriminator to maximize the sum of the probabilities over all the real classes. 

\textbf{Fake data}: When we obtain a generated example from the generator, we train the discriminator to classify it as a fake example.

Miyato et al. \cite{miyato2018virtual} proposed virtual adversarial training (VAT): a regularization method for both supervised and semi-supervised learning. Dai et al. \cite{dai2017good} show that given the discriminator objective, good semi-supervised learning indeed requires a bad generator from the theory perspective, and propose the definition of a preferred generator. A triangle GAN ($\Delta$-GAN) \cite{gan2017triangle} is proposed for semi-supervised cross-domain joint distribution matching and $\Delta$-GAN is closely related to Triple-GAN \cite{chongxuan2017triple}. Madani et al. \cite{madani2018semi} used semi-supervised learning with GANs for chest X-ray classification. 

Future improvements to GANs can be expected to simultaneously produce further improvements to semi-supervised learning and unsupervised learning such as self-supervised learning \cite{chen2019self}.

\subsubsection{Transfer learning}
Ganin et al. \cite{ganin2016domain} introduce a domain-adversarial training approach of neural networks for domain adaptation, where training data and test data are from similar but different distributions. The Professor Forcing algorithm \cite{lamb2016professor} uses adversarial domain adaptation for training recurrent network. Shrivastava et al. \cite{shrivastava2017learning} used GANs for simulated training data. A novel extension of pixel-level domain adaptation named GraspGAN \cite{bousmalis2018using} was proposed for robotic grasping     \cite{james2019sim,pinto2017supervision}. By using synthetic data and domain adaptation \cite{bousmalis2018using}, the number of real-world examples needed to achieve a given level of performance is reduced by up to 50 times, utilizing only randomly generated simulated objects.

Recent studies have shown remarkable success in image-to-image translation \cite{anoosheh2018combogan,chen2018attention,wang2018perceptual,cao2019biphasic,amodio2019travelgan,liu2019few} for two domains. However, existing methods such as CycleGAN \cite{zhu2017unpaired}, DiscoGAN \cite{kim2017learning}, and DualGAN \cite{yi2017dualgan}, cannot be used directly for more than two domains, since different approaches should be built independently for every pair of domains. StarGAN \cite{choi2018stargan} well solve this problem which can conduct image-to-image translations for multiple domains using only a single model. Other related works can be found in \cite{he2019attgan,liu2019stgan}. CoGAN \cite{liu2016coupled} can be also used for multiple domains.

Learning fair representations is a closely related problem to domain transfer. Note that different formulations of adversarial objectives \cite{edwards2015censoring,beutel2017data,zhang2018mitigating,madras2018learning} achieve different notations of fairness. 

Domain adaptation \cite{bousmalis2017unsupervised,hu2018duplex} can be seen as a subset of transfer learning \cite{pan2009survey}. Recent \textbf{visual domain adaptation (VDA)} methods include: \textbf{visual appearance adaptation, representation adaptation}, and \textbf{output adaptation}, which can be thought of as using domain adaptation based on the original input, features, and outputs of the domains, respectively.

\textbf{Visual appearance adaptation:} CycleGAN \cite{zhu2017unpaired} is a representative method in this category. CyCADA \cite{hoffman2018cycada} is proposed for visual appearance adaptation based on CycleGAN. 

\textbf{Representation adaptation:} The key of adversarial discriminative domain adaptation (ADDA) \cite{tzeng2017adversarial,wang2017catgan} is to learn feature representations that a discriminator cannot differentiate which domain they belong to. Sankaranarayanan et al. \cite{sankaranarayanan2018learning} focused on adapting the representations learned by segmentation networks across real and synthetic domains based on GANs. Fully convolutional adaptation networks (FCAN) \cite{zhang2018fully} is proposed for semantic segmentation which combines visual appearance adaptation and representation adaptation.

\textbf{Output adaptation:} Tsai \cite{tsai2018learning} made the outputs of the source and target images have a similar structure so that the discriminator cannot differentiate them. 

Other transfer learning based GANs can be found in \cite{ajakan2014domain,shen2017wasserstein,benaim2017one,zhao2018adversarial,long2018conditional,hong2018conditional,saito2018maximum,volpi2018adversarial}.

\subsubsection{Reinforcement learning}
Generative models can be integrated into reinforcement learning (RL) \cite{silver2016mastering} in different ways \cite{goodfellow2016nips,chen2019generative}. Reference \cite{pfau2016connecting} has already discussed connections between GANs and actor-critic methods. The connections among GANs, inverse reinforcement learning (IRL), and energy-based models is studied in \cite{finn2016connection}. These connections to RL are possible to be useful for the development of both GANs and RL. Furthermore, GANs were combined with reinforcement learning for synthesizing programs for images \cite{ganin2018synthesizing}. The competitive multi-agent learning framework proposed in \cite{bansal2017emergent} is also related to GANs and works on learning robust grasping policies by an adversary.

\textbf{Imitation Learning:} The connection between imitation learning and EBGAN is discussed in \cite{yoo2017energy}. Ho and Ermon \cite{ho2016generative} show that an instantiation of their framework draws an analogy between GANs and imitation learning, from which they derive a model-free imitation learning method that has significant performance gains over existing model-free algorithms in imitating complex behaviors in large and high-dimensional environments. Song et al. \cite{song2018multi} proposed multi-agent generative adversarial imitation learning (GAIL) and Guo et al. \cite{guo2018generative} proposed generative adversarial self-imitation learning. A multi-agent GAIL framework is used in a deconfounded multi-agent environment reconstruction (DEMER) approach \cite{shang2019environment} to learn the environment. DEMER is tested in the real application of Didi Chuxing and has achieved good performances.

\subsubsection{Multi-modal learning}
\begin{figure}
\centering
\scalebox{0.6}{\includegraphics{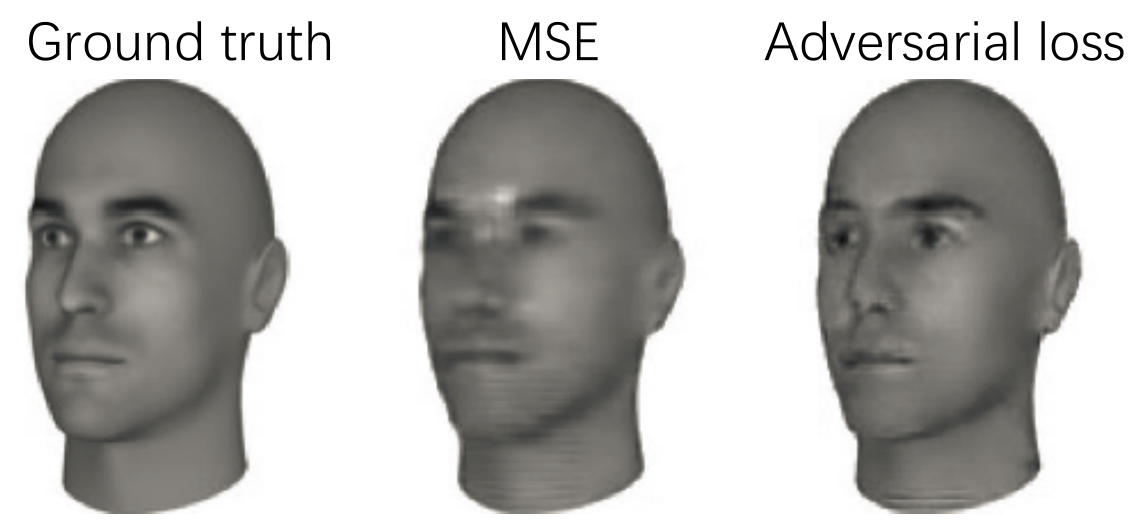}}
\caption{Lotter et al. \cite{lotter2015unsupervised} show an excellent description of the significance of being capable of modeling multi-modal data. In this instance, a model is trained to predict the next frame in a video. The video describes a computer rendering of a moving 3D model of a man's head. The image on the left is the ground truth, an instance of an actual frame of a video, which the model would like to predict. The image in the middle is what happens when the model is trained using mean squared error (MSE) between the model's predicted next frame and the actual next frame. The model is forced to select only one answer for what the next frame will be. Since there are multiple possible answers, corresponding to slightly diverse positions of the head, the answer that the model selects is an average over multiple slightly diverse images. This causes the faces to have a blurring effect. Utilizing an additional GANs loss, the image on the right is capable of knowing that there are multiple possible outputs, each of which is recognizable and clear as a realistic, satisfying image. Images are from \cite{lotter2015unsupervised}.}
\label{Fig:5}
\end{figure}

Generative models, especially GANs, make machine learning be able to work with multi-modal outputs. In many tasks, an input may correspond to multiple diverse correct outputs, each of which is an acceptable answer. Traditional ways of training machine learning methods, such as minimizing the mean squared error (MSE) between the model's predicted output and a desired output, are not capable of training models that can produce many different correct outputs. One instance of such a case is predicting the next frame in a video sequence, as shown in Fig. \ref{Fig:5}. Multi-modal image-to-image translation related works can be found in \cite{zhu2017toward,almahairi2018augmented,huang2018multimodal,lee2018diverse}.

\subsubsection{Other task driven GANs}
GANs have been used for feature learning such as feature selection \cite{jordon2018knockoffgan}, hashing \cite{song2018binary,zhang2018unsupervised,zhang2018sch,ghasedi2018unsupervised,wang2018semi,qiu2017deep,zhang2019generative,wang2019wegan}, and metric learning \cite{dai2017metric}.

MisGAN \cite{li2019misgan} was proposed to learn from incomplete data with GANs. Evolutionary GANs are proposed in \cite{wang2019evolutionary}. Ponce et al. \cite{ponce2019evolving} combined GANs and genetic algorithms to evolve images for visual neurons. GANs have also been used in other machine learning tasks \cite{wang2018multiagent} such as active learning \cite{zhu2017generative,xie2019learning}, online learning \cite{grnarova2017online}, ensemble learning \cite{tolstikhin2017adagan}, zero-shot learning \cite{zhu2018generative,Qin2020generative}, and multi-task learning \cite{yang2019task}.

\section{Theory}
In this section, we first introduce maximum likelihood estimation. Then, we introduce mode collapse. Finally, other theoretical issues such as inverse mapping and memorization are discussed.

\subsection{Maximum likelihood estimation (MLE)}
Not all generative models use maximum likelihood estimation (MLE). Some generative models do not utilize MLE, but can be made to do so (GANs belong to this category). It can be simply proved that minimizing the Kullback-Leibler Divergence (KLD) between ${{p_{data}(x)}}$ and ${p_g(x)}$ is equal to maximizing the log likelihood as the number of samples $m$ increases:
\begin{eqnarray}\label{equ:MLE1}
\begin{array}{l}
 {\theta ^*} = \mathop {\arg \;\min }\limits_\theta  \;KLD\left( {\left. {{p_{data}}} \right\|{p_g}} \right) \\ 
  = \mathop {\arg \;\min }\limits_\theta  \; - \int {{p_{data}}\left( x \right)\log \frac{{{p_g}\left( x \right)}}{{{p_{data}}\left( x \right)}}} dx \\ 
  = \mathop {\arg \;\min }\limits_\theta  \int {{p_{data}}\left( x \right)} \log {p_{data}}\left( x \right)dx \\ 
 \quad  - \int {{p_{data}}\left( x \right)} \log {p_g}\left( x \right)dx \\ 
  = \mathop {\arg \;\max }\limits_\theta  \int {{p_{data}}\left( x \right)} \log {p_g}\left( x \right)dx \\ 
  = \mathop {\arg \;\max }\limits_\theta  \;\mathop {\lim }\limits_{m \to \infty } \frac{1}{m}\sum\nolimits_{i = 1}^m {\log {p_g}\left( {{x_i}} \right)}.  \\ 
 \end{array}
 \end{eqnarray}
 
The model probability distribution ${p_\theta }\left( x \right)$ is replaced with ${p_g}\left( x \right)$ for notation consistency. Refer to Chapter 5 of \cite{goodfellow2016deep} for more information on MLE and other statistical estimators.

\subsection{Mode collapse}\label{subsection:mode}
GANs are hard to train, and it has been observed \cite{salimans2016improved,che2017mode} that they often suffer from mode collapse \cite{srivastava2017veegan,bau2019seeing}, in which the generator learns to generate samples from only a few modes of the data distribution but misses many other modes, even if samples from the missing modes exist throughout the training data. In the
worst case, the generator produces simply a single sample (complete collapse) \cite{Arjovsky2017towards,arora2017generalization}. In this subsection, we will first introduce two viewpoints of GANs mode collapse. Then, we will introduce methods that propose new objective functions or new structures to solve mode collapse.

\subsubsection{Two viewpoints: divergence and algorithmic}
We can resolve and understand GANs mode collapse and instability from two viewpoints: divergence and algorithmic. 

\textbf{Divergence viewpoint:} Roth et al. \cite{roth2017stabilizing} stabilizd training of GANs and their variants such as $f$-divergence based GANs ($f$-GAN) through regularization.

\textbf{Algorithmic viewpoint:} The numerics of common algorithms for training GANs are analyzed and a new algorithm that has better convergence properties is proposed in \cite{mescheder2017numerics}. Mescheder et al. \cite{mescheder2018training} showed that which training methods for GANs do actually converge.

\subsubsection{Methods overcoming mode collapse}
\textbf{Objective function based methods:} Deep regret analytic GAN (DRAGAN) \cite{kodali2017convergence} suggests that the mode collapse exists due to the occurence of a fake local Nash equilibrium in the nonconvex problem. DRAGAN solves this problem by constraining gradients of the discriminator around the real data manifold. It adds a gradient penalizing term which biases the discriminator to have a gradient norm of 1
around the real data manifold. Other methods such as EBGAN and Unrolled GAN (detailed in Section \ref{subsection:GANsTraining}) also belong to this category.

\textbf{Structure based methods:} Representative methods in this category include such as MAD-GAN \cite{ghosh2018multi} and MRGAN \cite{che2017mode} (detailed in Section \ref{subsection:GANsTraining})
 
There are also other methods to reduce mode collapse in GANs. For example, PACGAN \cite{lin2018pacgan} eases the pain of mode collapse by changing input to the discriminator.

\subsection{Other theoretical issues}
\subsubsection{Do GANs actually learn the distribution?} 
References \cite{arora2017generalization,arora2018gans,dumoulin2016adversarially} have both empirically and theoretically brought the concern to light that distributions learned by GANs suffer from mode collapse. In contrast, Bai et al. \cite{bai2018approximability} show that GANs can in principle learn distributions in Wasserstein distance (or KL-divergence in many situations) with polynomial sample complexity, if the discriminator class has strong discriminating power against the particular generator class (instead of against all possible generators). Liang et al. \cite{liang2018well} studied how well GANs learn densities, including nonparametric and parametric target distributions. Singh et al. \cite{singh2018nonparametric} further studied nonparametric density estimation
with adversarial losses.

\subsubsection{Divergence/Distance} 
Arora et al. \cite{arora2017generalization} show that training of GAN may not have good generalization properties; e.g., training may look successful but the generated distribution may be far from real data distribution in standard metrics. The popular distances such as Wasserstein and Jensen-Shannon (JS) may not generalize. However, generalization does occur by introducing a novel notion of distance between distributions, the neural net distance. Are there other useful divergences?

\subsubsection{Inverse mapping} 
GANs cannot learn the inverse mapping - projecting data back into the latent space. BiGANs \cite{donahue2016adversarial} (detailed in Section \ref{paragraph:autoencoders}) is proposed as a way of learning this inverse mapping. Dumoulin et al. \cite{dumoulin2016adversarially} introduce the adversarially learned inference (ALI) model (detailed in Section \ref{paragraph:autoencoders}), which jointly learns an inference network and a generation network utilizing an adversarial process. Arora et al. \cite{arora2017theoretical} show the theoretical limitations of Encoder-Decoder GAN architectures such as BiGANs \cite{donahue2016adversarial} and ALI \cite{dumoulin2016adversarially}. Creswell et al. \cite{creswell2019inverting} invert the generator of GANs. 

\subsubsection{Mathematical perspective such as optimization} 
Mohamed et al. \cite{mohamed2016learning} frame GANs within the algorithms for learning in implicit generative models that only specify a stochastic procedure with which to generate data. Gidel et al. \cite{gidel2018variational} looked at optimization approaches designed for GANs and casted GANs optimization problems in the general variational inequality framework. The convergence and robustness of training
GANs with regularized optimal transport is disscussed in \cite{sanjabi2018convergence}.

\subsubsection{Memorization} 
As for ``memorization of GANs'', Nagarajan et al. \cite{Nagarajan2018Theoretical} argue that making the generator ``learn to memorize'' the training data is a more difficult task than making it ``learn to output realistic but unseen data''.

\section{Applications}\label{section:Applications}
As discussed earlier, GANs are a powerful generative model which can generate realistic-looking samples with a random vector $z$. We neither need to know an explicit true data distribution nor have any mathematical assumptions. These advantages allow GANs to be widely applied to many areas such as image processing and computer vision, sequential data.

\subsection{Image processing and computer vision}
The most successful applications of GANs are in image processing and computer vision, such as image super-resolution, image synthesis and manipulation, and video processing.

\subsubsection{Super-resolution (SR)}
SRGAN \cite{ledig2017photo}, GANs for SR, is the first framework able to infer photo-realistic natural images for   upscaling factors. To further improve the visual quality of SRGAN, Wang et al. \cite{wang2018esrgan} thoroughly study three key components of SRGAN and improve each of them to derive an Enhanced SRGAN (ESRGAN). For example, ESRGAN uses the idea from relativistic GANs \cite{jolicoeur2018relativistic} to have the discriminator predict relative realness rather than the absolute value. Benefiting from these improvements, ESRGAN won the first place in the PIRM2018-SR Challenge (region 3) \cite{blau20182018} and got the best perceptual index. Based on CycleGAN \cite{zhu2017unpaired}, the Cycle-in-Cycle GANs \cite{yuan2018unsupervised} is proposed for unsupervised image SR. SRDGAN \cite{guan2019srdgan} is proposed to learn the noise prior for SR with DualGAN \cite{yi2017dualgan}. Deep tensor generative adversarial nets (TGAN) \cite{ding2019tgan} is proposed to generate large high-quality images by exploring tensor structures. There are methods specific for face SR \cite{yu2016ultra,zhu2018high,huang2019wavelet}. Other related methods can be found in \cite{sonderby2016amortised,johnson2016perceptual,wang2018recovering,Zhang2019RankSRGAN}.

\subsubsection{Image synthesis and manipulation}
\paragraph{Face}~{}\newline
\textbf{Pose related:} Disentangled representation learning GAN (DR-GAN) \cite{tran2017disentangled} is proposed for pose-invariant face recognition. Huang et al. \cite{huang2017beyond} proposed a Two-Pathway GAN (TP-GAN) for photorealistic frontal view synthesis by simultaneously perceiving local details and global structures. Ma et al. \cite{ma2017pose} proposed the novel Pose Guided Person Generation Network (PG$^2$) that synthesizes person images in arbitrary poses, based on a novel pose and an image of that person. Cao et al. \cite{cao2018learning} proposed a high fidelity pose invariant model for high-resolution face frontalization based on GANs. Siarohin et al. \cite{siarohin2018deformable} proposed deformable gans for pose-based human image generation. Pose-robust spatial-aware GAN (PSGAN) for customizable makeup transfer is proposed in \cite{Jiang2019PSGAN}.

\begin{figure}
\centering
\subfloat[photo]{\scalebox{0.468}{\includegraphics{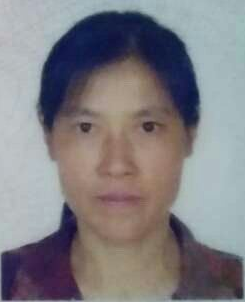}}}
\subfloat[portrait drawings]{\scalebox{0.108}{\includegraphics{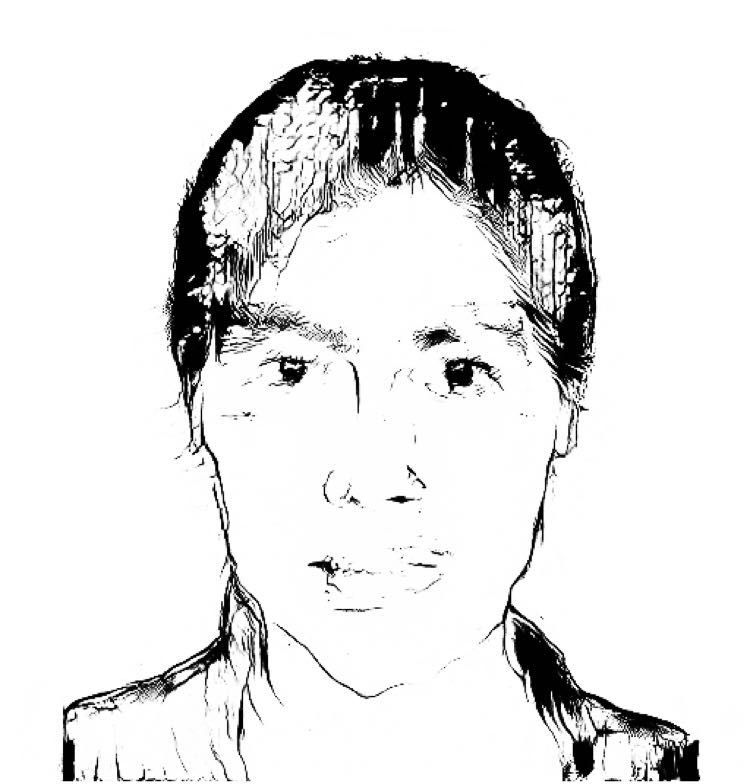}}}
\caption{Given a photo such as (a), APDrawingGAN can produce the corresponding artistic portrait drawings (b).}
\label{Fig:7}
\end{figure}

\textbf{Portrait related:} APDrawingGAN \cite{yi2019apdrawinggan} is proposed to generate artistic portrait drawings from face photos with hierarchical GANs. APDrawingGAN has a software based on wechat and the results are shown in Fig. \ref{Fig:7}. GANs have also been used in other face related applications such as facial attribute changes \cite{wang2017tag} and portrait editing \cite{shu2017neural,chang2018pairedcyclegan,dolhansky2018eye,pumarola2018ganimation}.

\begin{figure*}
\centering
\scalebox{0.52}{\includegraphics{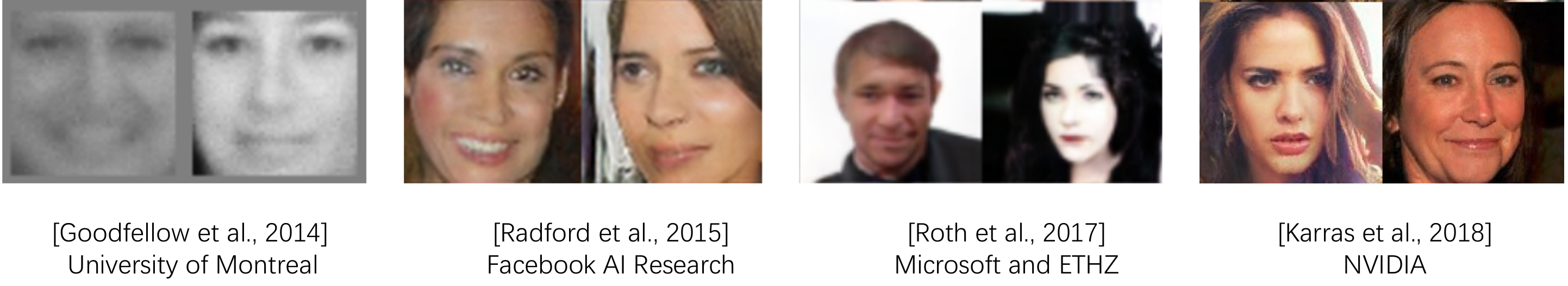}}\\
\caption{Face image synthesis.}
\label{Fig:6}
\end{figure*}

\textbf{Face generation:} The quality of generated faces by GANs is improved year by year, which can be found in Sebastian Nowozin's GAN lecture materials\footnote{\url{https://github.com/nowozin/mlss2018-madrid-gan}}. As we can see from Figure \ref{Fig:6}, the generated faces based on original GANs \cite{goodfellow2014generative} are of low visual quality and can only serves as a proof of concept. Radford et al. \cite{radford2015unsupervised} used better neural network architectures: deep convolutional neural networks for generating faces. Roth et al. \cite{roth2017stabilizing} addressed the instability problems of GAN training, allowing for larger architectures such as the ResNet to be utilized. Karras et al. \cite{karras2017progressive} utilized multiscale training, allowing megapixel face image generation at high fidelity. 

Face generation \cite{yin2017semi,donahue2017semantically,duarte2019wav2pix,gecer2019ganfit,shu2018deforming,fu2019dual,lu2018attribute,cao20193d,liu2019attribute} is somewhat easy because there is only one class of object. Every object is a face and most face data sets tend to be composed of people looking straight into the camera. Most people have been registered in terms of putting nose and eyes and other landmarks in consistent locations. 

\paragraph{General object}~{}\newline
It is a little harder to have GANs work on assorted data sets like ImageNet \cite{deng2009imagenet} which has a thousand different object classes. However, we have seen rapid progress over the recent few years. The quality of these images has been improved year by year \cite{mescheder2018training}.

While most papers use GANs to synthesize
images in two dimensions\cite{bao2017cvae,dong2017semantic}, Wu et al. \cite{wu2016learning} synthesized three-dimensional (3-D) samples using GANs and volumetric convolutions. Wu et al. \cite{wu2016learning} synthesized novel objects
including cars, chairs, sofa, and tables. Im et al. \cite{im2016generating} generated images with recurrent adversarial networks. Yang et al. \cite{yang2017lr} proposed layered recursive GANs (LR-GAN) for image generation.

\paragraph{Interaction between a human being and an image generation process}~{}\newline
There are many applications that involve interaction between a human being and an image generation process. Realistic image manipulation is difficult because it requires modifying the image in a user-controlled way, while making it appear realistic. If the user does not have efficient artistic skill, it is easy to deviate from the manifold of natural images while editing. Interactive GAN (IGAN) \cite{zhu2016generative} defines a class of image editing operations, and constrain their output to lie on that learned manifold at all times. Introspective adversarial networks \cite{brock2016neural} also offer this capability to perform interactive photo editing and have demonstrated their results mostly in face editing. GauGAN \cite{park2019semantic} can turn doodles into stunning, photorealistic landscapes.

\subsubsection{Texture synthesis}
Texture synthesis is a classical problem in image field. Markovian GANs (MGAN) \cite{li2016precomputed} is a texture synthesis method based on GANs. By capturing the texture data of Markovian patches, MGAN can generate stylized videos and images very quickly, in order to realize real-time texture synthesis. Spatial GAN (SGAN) \cite{jetchev2016texture} was the first to apply GANs with fully unsupervised learning in texture synthesis. Periodic spatial GAN (PSGAN) \cite{bergmann2017learning} is a variant of SGAN, which can learn periodic textures from a single image or complicated big dataset.

\subsubsection{Object detection}
How can we learn an object detector that is invariant to deformations and occlusions? One way is using a data-driven strategy - collect large-scale datasets which have object examples under different conditions. We hope that the final classifier can use these instances to learn invariances. Is it possible to see all the deformations and occlusions in a dataset? Some deformations and occlusions are so rare that they hardly happen in practical applications; yet we want to learn a method invariant to such situations. Wang et al. \cite{wang2017fast} used GANs to generate instances with deformations and occlusions. The aim of the adversary is to generate instances that are difficult for the object detector to classify. By using a segmentor and GANs, Segan \cite{ehsani2018segan} detected objects occluded by other objects in an image. To deal with the small object detection problem, Li et al. \cite{li2017perceptual} proposed perceptual GAN and Bai et al. \cite{bai2018sod} proposed an end-to-end multi-task GAN (MTGAN).

\subsubsection{Video applications}
Reference \cite{vondrick2016generating} is the first paper to use GANs for video generation. Villegas et al. \cite{villegas2017decomposing} proposed a deep neural network for the prediction of future frames in natural video sequences using GANs. Denton and Birodkar \cite{denton2017unsupervised} proposed a new model named disentangled representation net (DRNET) that learns disentangled image representations from video based on GANs. A novel video-to-video synthesis approach (video2video) under the generative adversarial learning framework was proposed in \cite{wang2018video}. MoCoGan \cite{tulyakov2018mocogan} is proposed to decompose motion and content for video generation \cite{santana2016learning,chan2018everybody,clark2019efficient}. 

GANs have also been used in other video applications such as
video prediction \cite{mathieu2015deep,walker2017pose,liang2017dual} and video retargeting \cite{bansal2018recycle}.

\subsubsection{Other image and vision applications}
GANs have been utilized in other image processing and computer vision tasks \cite{liang2018generative,kim2017unsupervised,chen2018cartoongan,villegas2018neural} such as object transfiguration \cite{zhou2017genegan,wu2017gp}, semantic segmentation \cite{souly2017semi}, visual saliency prediction \cite{pan2017salgan}, object tracking \cite{song2018vital,han2019robust}, image dehazing \cite{engin2018cycle,yang2018towards,liu2019end}, natural image matting \cite{lutz2018alphagan}, image inpainting \cite{yeh2017semantic, yu2018generative}, image fusion \cite{liu2018psgan}, image completion \cite{iizuka2017globally}, and image classification \cite{liu2019task}.

Creswell et al. \cite{creswell2016adversarial} show that the representations learned by GANs can also be used for retrieval. GANs have also been used for anticipating where people will look \cite{zhang2017deep,zhang2018anticipating}.

\subsection{Sequential data}
GANs also have achievements in sequential data such as natural language, music, speech, voice \cite{fang2018high,kaneko2017parallel}, and time series \cite{esteban2017real,hartmann2018eeg,donahue2018synthesizing,li2019mad}.

\textbf{Natural language processing (NLP): } IRGAN \cite{wang2017irgan,lu2019psgan} is proposed for information retrieval (IR). Li et al. \cite{li2017adversarial} used adversarial learning for neural dialogue generation. GANs have also been used in text generation \cite{zhang2016generating,lin2017adversarial,fedus2018maskgan,yang2019tet} and speech language processing \cite{yu2017seqgan}. Kbgan \cite{cai2017kbgan} is proposed to generate high-quality negative examples and used in knowledge graph embeddings. Adversarial REward Learning (AREL) \cite{wang2018no} is proposed for visual storytelling. DSGAN \cite{qin2018dsgan} is proposed for distant supervision relation extraction. ScratchGAN \cite{d2019training} is proposed to train a language GAN from scratch -- without maximum likelihood pre-training. 

Qiao et al. \cite{qiao2019mirrorgan} learn text-to-image generation by redescription and text conditioned auxiliary classifier GAN (TAC-GAN) \cite{dash2017tac} is also proposed for text to image. GANs have been widely used in image-to-text (image caption) \cite{chen2017show,shetty2017speaking}, too.

Furthermore, GANs have been widely utilized in other NLP applications such as question answer selection \cite{rao2019answer,yang2019adversarial}, poetry generation \cite{liu2018beyond}, talent-job fit \cite{luo2019resumegan}, and review detection and geneneration \cite{garbacea2019judge,aghakhani2018detecting}.

\textbf{Music: }GANs have been used for generating music such as continuous RNN-GAN (C-RNN-GAN) \cite{mogren2016c}, Object-Reinforced GAN (ORGAN) \cite{guimaraes2017objective}, and SeqGAN \cite{lee2017seqgan,yu2017seqgan}.

\textbf{Speech and Audio: }GANs have been used for speech and audio analysis such as synthesis \cite{takuhiro2017generative,saito2018statistical,donahue2018adversarial}, enhancement \cite{pascual2017segan}, and recognition \cite{donahue2018exploring}, .

\subsection{Other applications}
\textbf{Medical field: }GANs have been widely utilized in medical field such as generating and designing DNA \cite{killoran2017generating,gupta2018feedback}, drug discovery \cite{benhenda2017chemgan}, generating multi-label discrete patient records \cite{choi2017generating}, medical image processing \cite{dai2017scan,schlegl2017unsupervised,wolterink2017deep,quan2018compressed,mardani2018deep,xue2018segan,yang2018low,st2018generative}, dental restorations \cite{hwang2018learning}, and doctor recommendation \cite{tian2019drgan}.

\textbf{Data science: }GANs have been used in data generating \cite{zheng2017unlabeled,gorijala2017image,wang2017adversarial,chang2018generating,sixt2018rendergan,xu2018fairgan,lee2018rare,xu2019CFGAN,turkoglu2019layer,eltell}, neural networks generating \cite{ratzlaff2019hypergan}, data augmentation \cite{frid2018gan, wang2019enhancing}, spatial representation learning \cite{zhang2019unifying}, network embedding \cite{gao2019progan}, heterogeneous information networks \cite{hu2019adversarial},
and mobile user profiling \cite{wang2019adversarial}.

GANs have been widely used in many other areas such as malware detection \cite{hu2017generating}, chess game playing \cite{chidambaram2017style}, steganography \cite{chu2017cyclegan,volkhonskiy2017steganographic,shi2017ssgan,hayes2017generating}, privacy-preserving \cite{abadi2016learning,gomez2018unsupervised,beaulieu2018privacy}, social robot \cite{gupta2018social}, and network pruning \cite{shu2019co,lin2019towards}.

\section{Open research problems}
Because GANs have become popular throughout the
deep learning area, its limitations have recently been improved \cite{daskalakis2017training,bora2018ambientgan}. There are still open research problems for GANs.

\textbf{GANs for discrete data:} GANs rely on the generated samples being completely differentiable with respect to the generative parameters. Therefore, GANs cannot produce discrete data directly, such as hashing code and one-hot word. Solving this problem is very important since it could unlock the potential of GANs for NLP and hashing. Goodfellow \cite{goodfellow2016nips} suggested three ways to solve this problem: using Gumbel-softmax \cite{kusner2016gans,jang2016categorical} or the concrete distribution \cite{maddison2016concrete}; utilizing the REINFORCE algorithm \cite{williams1992simple}; training the generator to sample continuous values that can be transformed to discrete ones (such as sampling word embeddings directly).

There are other methods towards this research direction. Song et al. \cite{song2018binary} used a continuous function to approximate the sign function for hashing code. Gulrajani et al. \cite{gulrajani2017improved} modelled discrete data with a continuous generator. Hjelm et al. \cite{hjelm2018boundary} introduced an algorithm for training GANs with discrete data that utilizes the estimated difference measure from the discriminator to compute importance weights for generated samples, and thus providing a policy gradient for training the generator. Other related work can be found in \cite{che2017maximum,junbo2017adversarially}. More work needs to be done in this interesting area.


\textbf{New Divergences:} New families of Integral Probability Metrics (IPMs) for training GANs such as Fisher GAN \cite{mroueh2017fisher,yoo2017domain}, mean and covariance feature matching GAN (McGan) \cite{mroueh2017mcgan}, and Sobolev GAN \cite{mroueh2017sobolev}, have been proposed. Are there any other interesting classes of divergences? This deserves further study.

\textbf{Estimation uncertainty:} Generally speaking, as we have more data, uncertainty estimation reduces. GANs do not give the distribution that generated the training examples and GANs aim to generate new samples that come from the same distribution of the training examples. Therefore, GANs have neither a likelihood nor a well-defined posterior. There are early attempts towards this research direction such as Bayesian GAN \cite{saatci2017bayesian}. Although we can use GANs to generate data, how can we measure the uncertainty of the well-trained generator? This is another interesting future issue.

\textbf{Theory:} As for generalization, Zhang et al. \cite{zhang2017discrimination} developed generalization bounds between the true distribution and learned distribution under different evaluation metrics. When evaluated with neural distance, the bounds in \cite{zhang2017discrimination} show that generalization is guaranteed as long as the discriminator set is small enough, regardless of the size of the hypothesis set or generator. Arora et al. \cite{arora2018gans} proposed a novel test for estimating support size using the birthday paradox of discrete probability and show that GAN does suffer mode collapse even when images are of higher visual quality. More deep theoretical research is well worth studying. How do we test for generalization empirically? Useful theory should enable choice of model class, capacity, and architectures. This is an interesting issue to be investigated in future work.

\textbf{Others:} There are other important research problems for GANs such as evaluation (detailed in Subsection \ref{subsection:How}) and mode collapse (detailed in Subsection \ref{subsection:mode})

\section{Conclusions}
This paper presents a comprehensive review of
various aspects of GANs. We elaborate on several perspectives, i.e., algorithm, theory, applications, and open research problems. We believe this
survey will help readers to gain a thorough understanding of the GANs research area.


%

\appendices


\ifCLASSOPTIONcompsoc
  \section*{Acknowledgments}
\else
  \section*{Acknowledgment}
\fi

The authors would like to thank the NetEase course taught by Shuang Yang, Ian Good fellow's invited talk at AAAI 19, CVPR 2018 tutorial on GANs, Sebastian Nowozin's MLSS 2018 GAN lecture materials. The authors also would like to thank the helpful discussions with group members of Umich Yelab and Foreseer research group.

\ifCLASSOPTIONcaptionsoff
  \newpage
\fi

\bibliographystyle{ieeetr}
\bibliography{a}








\end{document}